\documentclass[sigconf]{acmart}
\AtBeginDocument{%
  \providecommand\BibTeX{{%
    \normalfont B\kern-0.5em{\scshape i\kern-0.25em b}\kern-0.8em\TeX}}}

\copyrightyear{2023}
\acmYear{2023}
\setcopyright{rightsretained}
\acmConference[KDD '23]{Proceedings of the 29th ACM SIGKDD Conference on Knowledge Discovery and Data Mining}{August 6--10, 2023}{Long Beach, CA, USA}
\acmBooktitle{Proceedings of the 29th ACM SIGKDD Conference on Knowledge Discovery and Data Mining (KDD '23), August 6--10, 2023, Long Beach, CA, USA}
\acmDOI{10.1145/3580305.3599861}
\acmISBN{979-8-4007-0103-0/23/08}

\makeatletter
\gdef\@copyrightpermission{
  \begin{minipage}{0.3\columnwidth}
   \href{https://creativecommons.org/licenses/by/4.0/}{\includegraphics[width=0.90\textwidth]{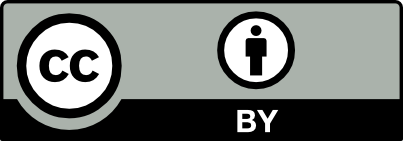}}
  \end{minipage}\hfill
  \begin{minipage}{0.7\columnwidth}
   \href{https://creativecommons.org/licenses/by/4.0/}{This work is licensed under a Creative Commons Attribution International 4.0 License.}
  \end{minipage}
  \vspace{5pt}
}
\makeatother

\usepackage{soul}
\usepackage{graphicx}

\usepackage{booktabs}
\usepackage{fancybox}
\usepackage{multirow}
\usepackage{subfigure}
\usepackage{verbatim}
\usepackage{sidecap,color}
\usepackage{microtype}
\usepackage{mathtools}
\usepackage{multirow, paralist, color}
\usepackage{amsmath,enumitem}

\usepackage{amsfonts}

\usepackage{subfigure}
\usepackage{booktabs} % for professional tables
\usepackage{textcase}
\usepackage[T1]{fontenc}
\usepackage{makecell}
\usepackage{listings}
\usepackage{setspace}
\usepackage{amsthm}
\usepackage{graphicx}

\newtheorem{definition}{Definition}

\usepackage{xcolor}
\usepackage[linesnumbered,ruled]{algorithm2e} %linesnumbered,ruled,vlined

\setlist[itemize]{noitemsep, topsep=0pt}

\makeatletter
\newcommand{\removelatexerror}{\let\@latex@error\@gobble}
\makeatother

\definecolor{babypink}{rgb}{0.96, 0.76, 0.76}
\definecolor{almond}{rgb}{0.94, 0.87, 0.8}
\definecolor{arylideyellow}{rgb}{0.91, 0.84, 0.42}
\definecolor{blond}{rgb}{0.98, 0.94, 0.75}
\definecolor{carolinablue}{rgb}{0.6, 0.73, 0.89}
\definecolor{columbiablue}{rgb}{0.61, 0.87, 1.0}
\definecolor{hollywoodcerise}{rgb}{0.96, 0.0, 0.63}
\definecolor{ashgrey}{rgb}{0.7, 0.75, 0.71}

\definecolor{codegreen}{rgb}{0.313, 0.498, 0.498} %{0,0.6,0}
\definecolor{codegray}{rgb}{0.5,0.5,0.5}
\definecolor{codepurple}{rgb}{0.58,0,0.82}
\definecolor{backcolour}{rgb}{0.95,0.95,0.92}

\lstdefinestyle{mystyle}{
    commentstyle=\color{codegreen},
    keywordstyle=\color{magenta},
    numberstyle=\tiny\color{codegray},
    stringstyle=\color{codepurple},
    basicstyle=\ttfamily\footnotesize,
    breakatwhitespace=false,         
    breaklines=true,                 
    captionpos=b,                    
    keepspaces=true,                 
    numbers=left,                    
    numbersep=5pt,                  
    showspaces=false,                
    showstringspaces=false,
    showtabs=false,                  
    tabsize=2
}
\begin{document}

\title{LibAUC: A Deep Learning Library for X-Risk Optimization}

%\author{Zhuoning Yuan}
%\email{zhuoning-yuan@uiowa.edu}
\author{Zhuoning Yuan}
\affiliation{%
  \institution{University of Iowa}
  %\streetaddress{P.O. Box 1212}
  \country{}
}
\email{zhuoning-yuan@uiowa.edu}

\author{Dixian Zhu}
\authornote{contributed equally to
this work.}
\affiliation{%
  \institution{University of Iowa}
  %\streetaddress{1 Th{\o}rv{\"a}ld Circle}
  %\city{Hekla}
  \country{}
  }
\email{dixian-zhu@uiowa.edu}

\author{Zi-Hao Qiu}
\authornotemark[1]
\affiliation{%
  \institution{Nanjing University}
  %\city{Rocquencourt}
  \country{}
}
\email{qiuzh@lamda.nju.edu.cn}

\author{Gang Li}
\authornotemark[1]
\affiliation{%
  \institution{University of Iowa}
 %\streetaddress{Rono-Hills}
 %\city{Doimukh}
 %\state{Arunachal Pradesh}
 \country{}
 }
\email{gang-li@uiowa.edu}

\author{Xuanhui Wang}
\affiliation{%
  \institution{Google Research}
  %\streetaddress{30 Shuangqing Rd}
  %\city{Haidian Qu}
  %\state{Beijing Shi}
  \country{}
  }
\email{xuanhui@google.com}

\author{Tianbao Yang}
\authornote{Correponding author}
\affiliation{
  \institution{Texas A\&M University}
  %\streetaddress{8600 Datapoint Drive}
  %\city{San Antonio}
  %\state{Texas}
  \country{}
  %\postcode{78229}
  }
\email{tianbao-yang@tamu.edu}

\def \textM {{\text{M}}}
\def \textS {{\text{S}}}

\def\true{{\text{true}}}
\def\EX{{\mathbb{E}}}

\def\bw{\mathbf{w}}
\def\mbI{\mathbb{I}}

\def \S {\mathbf{S}}
\def \A {\mathcal{A}}
\def \X {\mathcal{X}}
\def \Y {\mathcal{Y}}
\def \Ab {\bar{\A}}
\def \R {\mathbb{R}}
\def \Kt {\widetilde{K}}
\def \k {\mathbf{k}}
\def \w {\mathbf{w}}
\def \v {\mathbf{v}}
\def \t {\mathbf{t}}
\def \x {\mathbf{x}}
\def \Se {\mathcal{S}}
\def \E {\mathbb{E}}
\def \Rh {\widehat{R}}
\def \x {\mathbf{x}}
\def \p {\mathbf{p}}
\def \a {\mathbf{a}}
\def \diag {\mbox{diag}}
\def \b {\mathbf{b}}
\def \e {\mathbf{e}}
\def \ba {\boldsymbol{\alpha}}
\def \c {\mathbf{c}}
\def \tr {\mbox{tr}}
\def \d {\mathbf{d}}
\def \db {\bar\mathbf{d}}
\def \1 {\mathbf{1}}

\def \z {\mathbf{z}}
\def \s {\mathbf{s}}
\def \bh {\widehat{\b}}
\def \y {\mathbf{y}}
\def \uh {\widehat{\u}}
\def \H {\mathcal{H}}
\def \g {\mathbf{g}}
\def \F {\mathcal{F}}
\def \I {\mathbb{I}}
\def \P {\mathcal{P}}
\def \Q {\mathcal{Q}}
\def \xh {\widehat{\x}}
\def \wh {\widehat{\w}}
\def \ah {\widehat{\a}}
\def \Rc {\mathcal R}
\def \Sigmah {\widehat\Sigma}

\def \Bh {\widehat B}
\def \Ah {\widehat A}
\def \Uh {\widehat U}
\def \Ut {\widetilde U}
\def \B {\mathcalB}
\def \C {\mathbf C}
\def \U {\mathbf U}
\def \Kh {\widehat K}
\def \fh {\widehat f}
\def \yh {\widehat\y}
\def \Xh {\widehat{X}}
\def \Fh {\widehat{F}}

\def \y {\mathbf{y}}
\def \x {\mathbf{x}}
\def \g {\nabla{g}}
\def \D {\mathcal{D}}
\def \z {\mathbf{z}}
\def \u {\mathbf{u}}
\def \H {\mathcal{H}}
\def \Z {\mathcal{Z}}
\def \Pc {\mathcal{P}}
\def \w {\mathbf{w}}
\def \r {\mathbf{r}}
\def \R {\mathbb{R}}
\def \S {\mathcal{S}}
\def \regret {\mbox{regret}}
\def \Uh {\widehat{U}}
\def \Q {\mathcal{Q}}
\def \W {\mathcal{W}}
\def \N {\mathcal{N}}
\def \A {\mathcal{A}}
\def \q {\mathbf{q}}
\def \v {\mathbf{v}}
\def \M {\mathcal{M}}
\def \c {\mathbf{c}}
\def \ph {\widehat{p}}
\def \d {\mathbf{d}}
\def \p {\mathbf{p}}
\def \q {\mathbf{q}}
\def \db {\bar{\d}}
\def \dbb {\bar{d}}
\def \I {\mathbb{I}}
\def \xt {\widetilde{\x}}
\def \yt {\widetilde{\y}}

\def \f {\mathbf{f}}
\def \a {\mathbf{a}}
\def \b {\mathbf{b}}
\def \ft {\widetilde{\f}}
\def \bt {\widetilde{\b}}
\def \h {\mathbf{h}}
\def \B {\mathcal{B}}
\def \bts {\widetilde{b}}
\def \fts {\widetilde{f}}
\def \Gh {\widehat{G}}
\def \G {\mathcal {G}}
\def \bh {\widehat{b}}
\def \fh {\widehat{\f}}
\def \wh {\widehat{\w}}
\def \vb {\bar{\mathbf v}}
\def \zt {\widetilde{\z}}
\def \zh {\widehat{\z}}
\def \zts {\widetilde{z}}
\def \s {\mathbf{s}}
\def \gh {\widehat{\g}}
\def \vh {\widehat{\v}}
\def \Sh {\widehat{S}}
\def \rhoh {\widehat{\rho}}
\def \hh {\widehat{\h}}
\def \C {\mathcal{C}}
\def \V {\mathcal{L}}
\def \t {\mathbf{t}}
\def \xh {\widehat{\x}}
\def \Ut {\widetilde{U}}
\def \wt {\widetilde{\w}}
\def \Th {\widehat{T}}
\def \Ot {\tilde{\mathcal{O}}}
\def \X {\mathcal{X}}
\def \nb {\widehat{\nabla}}
\def \K {\mathcal{K}}
\def \P {\mathbb{P}}
\def \T {\mathcal{T}}
\def \F {\mathcal{F}}
\def \ft{\widetilde{f}}
\def \Rt {\mathcal{R}}
\def \Rb {\bar{\Rt}}
\def \wb {\bar{\w}}
\def \zu {\underline{\z}}

\definecolor{myblue}{HTML}{2677cc} % Define a custom color
\definecolor{myred}{HTML}{f83f3c} % Define a custom color

\renewcommand{\shortauthors}{Zhuoning Yuan et al.}
%% No italics and no comma
%% If needed use a foot or author note to identify equal contribution
\begin{abstract}
This paper introduces the award-winning deep learning (DL) library called {\it LibAUC} for implementing state-of-the-art algorithms towards optimizing a family of risk functions named {\it X-risks}. X-risks refer to a family of {\it compositional functions} in which the loss function of each data point is defined in a way that contrasts the data point with a large number of others. They have broad applications in AI for solving classical and emerging problems, including but not limited to classification for imbalanced data (CID), learning to rank (LTR), and contrastive learning of representations (CLR). The motivation of developing LibAUC is to address the convergence issues of existing libraries for solving these problems. In particular, existing libraries may not converge or require very large mini-batch sizes in order to attain good performance for these problems, due to the usage of the standard mini-batch technique in the empirical risk minimization (ERM) framework. Our library is for {\it deep X-risk optimization (DXO)} that has achieved great success in solving a variety of tasks for CID, LTR and CLR. The contributions of this paper include: (1) It introduces a new mini-batch based pipeline for implementing DXO algorithms, which differs from existing DL pipeline in the design of {\it controlled data samplers and dynamic mini-batch losses}; (2) It provides extensive benchmarking experiments for ablation studies and comparison with existing libraries. The LibAUC library features scalable performance for millions of items to be contrasted, faster and better convergence than existing libraries for optimizing X-risks, seamless PyTorch deployment  and versatile APIs for various loss optimization. Our library is available to the open source community at~\url{https://github.com/Optimization-AI/LibAUC}, to facilitate further academic research and industrial applications.
\end{abstract}

% \footnote{\url{https://github.com/Optimization-AI/LibAUC}}





\maketitle
\section{Introduction}
Deep learning (DL) platforms  such as TensorFlow~\cite{abadi2016TensorFlow} and PyTorch~\cite{paszke2019PyTorch} have dramatically reduced the efforts of developers and researchers for implementing different DL methods without worrying about low-level computations (e.g., automatic differentiation, tensor operations, etc). Based on these platforms, plenty of DL libraries have been developed for different purposes, which can be organized into different categories including (i) supporting specific tasks~\cite{pasumarthi2019tf,goyal2021vissl}, e.g., \texttt{TF-Ranking} for LTR~\cite{pasumarthi2019tf}, \texttt{VISSL} for self-supervised learning (SSL)~\cite{goyal2021vissl}, (ii) supporting specific data, e.g., \texttt{DGL} and \texttt{DIG} for graphs~\cite{wang2019dgl,JMLR:v22:21-0343};  (iii) supporting specific models~\cite{garden2020github,rw2019timm, wolf-etal-2020-transformers}, e.g., \texttt{Transformers} for transformer models~\cite{wolf-etal-2020-transformers}.   

However, it has been observed that these existing platforms and libraries have encountered some unique challenges when solving some classical and emerging problems in AI, including classification for imbalanced data (CID), learning to rank (LTR), contrastive learning of representations (CLR).  In particular, prior works have observed that large mini-batch sizes are necessary to attain good performance for these problems~\cite{brown2020smooth,Cakir_2019_CVPR,Rolinek_2020_CVPR,DBLP:journals/corr/abs-2108-11179,simclrv1,DBLP:conf/icml/RadfordKHRGASAM21}, which restricts the capabilities of these AI models in the real-world.  The reason for this issue is two-fold. First, the standard empirical risk minimization (ERM) framework,  which serves as the foundation of the standard mini-batch based methods, does not provide a good abstraction for many non-decomposable objectives in ML and ignores their inherent complexities.  Second, all existing DL libraries are developed based on the standard mini-batch based technique for ERM, which updates model parameters based on the gradient of a mini-batch loss as an approximation for the objective on the whole data set.

To address the first issue, a novel learning paradigm named deep X-risk optimization (DXO) was recently introduced~\cite{yang2022algorithmic}, which provides a unified framework to abstract the optimization of many compositional loss functions, including surrogate losses for AUROC, AUPRC/AP, and partial AUROC that are suitable for CID~\cite{yuan2021large,qi2021stochastic,zhu2022auc}, surrogate losses for NDCG, top-$K$ NDCG, and listwise losses that are used in LTR~\cite{qiu2022largescale}, and global contrastive losses for CLR~\cite{yuan2022provable}. To address the second issue, the LibAUC library implemented state-of-the-art algorithms for optimizing a variety of X-risks arising in CID, LTR and CLR. It has been used by many projects~\cite{10.1145/3485447.3512178,wei2021pooling,tecentyoutu,mxaihsieh,He_XI_Ebadi_Tremblay_Wong_2022,dang2022auc} and achieved great success in solving real-world problems, e.g., the 1st Place at the Stanford CheXpert Competition~\cite{yuan2021large} and MIT AICures Challenge~\cite{wang2021advanced}. Hence, it deserves in-depth discussions about the design principles and unique features  to facilitate future research and development for DXO. 

This paper aims to present the underlying design principles of the LibAUC library and provide a comprehensive study of the library regarding its unique features of design and superior performance compared to existing libraries.  The unique design features of the LibAUC library include (i) {\it dynamic mini-batch losses}, which are designed for computing the stochastic gradients of X-risks by automatic differentiation to ensure the convergence; (ii) {\it controlled data samplers}, which differ from standard random data samplers in that the ratio of the number of positive data to the number of negative data can be controlled and tuned to boost the performance. The superiority of the LibAUC library lies in: (i) it is scalable to millions of items to be ranked or contrasted with respect to an anchor data;   (ii) it is robust to small mini-batch sizes due to that all implemented algorithms have theoretical convergence guarantee regardless of mini-batch sizes; and (iii) it converges faster and to better solutions than existing libraries for optimizing a variety of compositional losses/measures suitable for CID, LTR and CLR.   

\setlength{\textfloatsep}{2pt}% Remove \textfloatsep

\setlength\abovedisplayskip{2pt}
\setlength\belowdisplayskip{2pt}

To the best of our knowledge, LibAUC is the first DL library that provides easy-to-use APIs  for optimizing a wide range of X-risks. Our main contributions for this work are summarized as follows:
\newline
\begin{itemize}
    \item We propose a novel DL pipeline to support efficient implementation of DXO algorithms, and provide implementation details of two unique features of our pipeline, namely dynamic mini-batch losses and controlled data samplers.
    \item We present extensive empirical studies to demonstrate the effectiveness of the unique features of the LibAUC library, and the superior performance of LibAUC compared to existing DL libraries/approaches for solving the three tasks, i.e.,  CID, LTR and CLR.
\end{itemize}

\vspace*{-0.1in}\section{Deep X-Risk Optimization (DXO)}\label{sec:dxo}
This section provides necessary background about DXO. We refer  readers to~\cite{yang2022algorithmic} for more discussions about theoretical guarantees. 

\vspace*{-0.1in}\subsection{A Brief History} The min-max optimization for deep AUROC maximization was studied in several earlier works~\cite{yuan2021large,liu2019stochastic}. Later, deep AUPRC/AP maximization was proposed by Qi et al.~\cite{qi2021stochastic}, which formulates the problem as a novel class of finite-sum coupled compositional optimization (FCCO) problem. The algorithm design and analysis for FCCO were improved in subsequent works~\cite{pmlr-v162-wang22ak,DBLP:conf/aistats/0006YZY22,DBLP:conf/nips/JiangLW0Y22}. Recently, the FCCO techniques were used for partial AUC maximization~\cite{zhu2022auc},  NDCG and top-$K$ NDCG optimization~\cite{qiu2022largescale}, and stochastic optimization of global contrastive losses with a small batch size~\cite{yuan2022provable}. More recently, Yang et al. \cite{yang2022algorithmic} proposed the  X-risk optimization framework, which aims to provide a unified venue for studying the optimization of different X-risks. The difference between this work and these previous works is that we aim to provide a technical justification for the library design towards implementing DXO algorithms for practical usage, and comprehensive studies of unique features and superiority of LibAUC over existing DL libraries.  

\subsection{Notations}
For \textbf{CID}, let $\S=\{(\x_1,y_1), \ldots, (\x_n, y_n)\}$ denote a set of training data, where $\x_i\in\mathcal X\subset\R^{d_{in}}$ denotes the input feature vector and $y_i\in\{1, -1\}$ denotes the corresponding label.  Let $\S_+=\{\x_i: y_i=1\}$ contain $n_+$ positive examples and $\S_-=\{\x_i: y_i=-1\}$ contain $n_-$ negative examples. Denote by $h_\w(\x):\mathcal X\rightarrow\R$ a parametric predictive function (e.g., a deep neural network) with a parameter $\w\in\R^d$. We use $\E_{\x\sim\S} = \frac{1}{|\S|}\sum_{\x\in\S}$ interchangeably below. 

For \textbf{LTR},  let $\Q$ denote a set of $N$ queries. For a query $q\in\Q$, let $\S_q=\{\x^q_i, i=1, \ldots, N_q\}$ denote a set of $N_q$ items (e.g., documents, movies) to be ranked. For each $\x^q_i\in\S_q$, let $y^q_i\in\R^+$ denote its relevance score, which measures the relevance between query $q$ and item $\x^q_i$. Let $\S^+_q\subseteq\S_q$ denote a set of $N^+_q$ (positive) items \emph{relevant} to $q$, whose relevance scores are \emph{non-zero}. Let $\S=\{(q, \x^q_i),q\in\Q, \x^q_i\in\S^+_q\}$ denote all relevant query-item (Q-I) pairs.  Denote by $h_\w(\x; q):\mathcal X\times\Q\rightarrow\R$ a parametric predictive function that outputs a predicted relevance score for $\x$ with respect to $q$.

For \textbf{CLR}, let $\S=\{\x_1,\ldots, \x_n\}$ denote a set of anchor data, and let $\S_i^{-}$ denote a set containing all negative samples with respect to $\x_i$. For unimodal SSL, $\S_i^-$ can be constructed by applying different data augmentations to all data excluding $\x_i$. For bimodal SSL, $\S_i^-$ can be constructed by including the different view of all data excluding $\x_i$. The goal of representation learning is to learn a feature encoder network $h_\w(\cdot)\in\R^{d_{\text{o}}}$ parameterized by a vector $\w\in\R^d$ that outputs an encoded feature vector for an input data .  
\subsection{The X-Risk Optimization Framework}
We use the following definition of X-risks given by~\cite{yang2022algorithmic}. 
\vspace*{-0.05in}    \begin{definition}(\cite{yang2022algorithmic})
       {\large\bf X-risks} refer to a family of compositional measures in which the loss function of each data point is defined in a way that contrasts the data point with a large number of others.
   % \vspace*{-0.1in}
Mathematically, X-risk optimization can be cast into the following abstract optimization problem: 
\begin{align}\label{eqn:xrisk} 
  \min_{\w\in\R^d}F(\w) = \frac{1}{|\S|}\sum\nolimits_{\z_i\in\S}f_i(g(\w; \z_i, \S_i)),
\end{align}
where $g:\R^d\mapsto\mathcal{R}$ is a mapping, $f_i:\mathcal{R} \mapsto \R$ is a simple deterministic function, $\S=\{\z_1, \ldots, \z_m\}$ denotes a target set of data points, and $\S_i$ denotes a reference set of data points dependent or independent of $\z_i$. 
    \end{definition}
The most common form of $g(\w; \z, \S)$ is the following: 
\begin{align*}
   g(\w; \z_i, \S_i) = \frac{1}{|\S_i|}\sum\nolimits_{\z_j\in\S_i}\ell(\w; \z_i, \z_j),
  %  &\text{The Optimization Form:} \quad \quad g(\w; \z_i, \S_i) =\arg\min_{\u\in\Omega} \frac{1}{|\S_i|}\sum_{\z_j\in\S_i}\ell(\u, \w;  \z_i, \z_j),
\end{align*}
where $\ell(\w; \z_i, \z_j) = \ell(h_\w(\z_i), h_\w(\z_j))$ is a pairwise loss. 

As a result, many DXO problems will be formulated as \textbf{FCCO}~\cite{pmlr-v162-wang22ak}:
\begin{align}\label{eqn:fcco}
    \min_{\w}\frac{1}{|\S|}\sum\nolimits_{\z_i\in\S}f_i\left(\frac{1}{|\S_i|}\sum\nolimits_{\z_j\in\S_i}\ell(h_\w(\z_i), h_\w(\z_j))\right). 
\end{align}
 %When $g(\w; \z_i, \S_i)=\frac{1}{|\S_i|}\sum_{\z'\in\S_i}\ell(\w; \z_i, \z')$,  problem (\ref{eqn:xrisk}) reduces to (\ref{eqn:fcco}). 
The FCCO problem is subtly different from the traditional stochastic compositional optimization~\cite{DBLP:journals/mp/WangFL17} due to the coupling of a pair of data in the inner function. Almost all X-risks considered in this paper, including AUROC, AUPRC/AP, pAUC, NDCG, top-$K$ NDCG, listwise CE loss, GCL, can be formulated as FCCO or its variants.

\begin{figure}[t]
%\vskip 0.2in
\begin{center}
\centerline{\includegraphics[width=0.9\columnwidth]{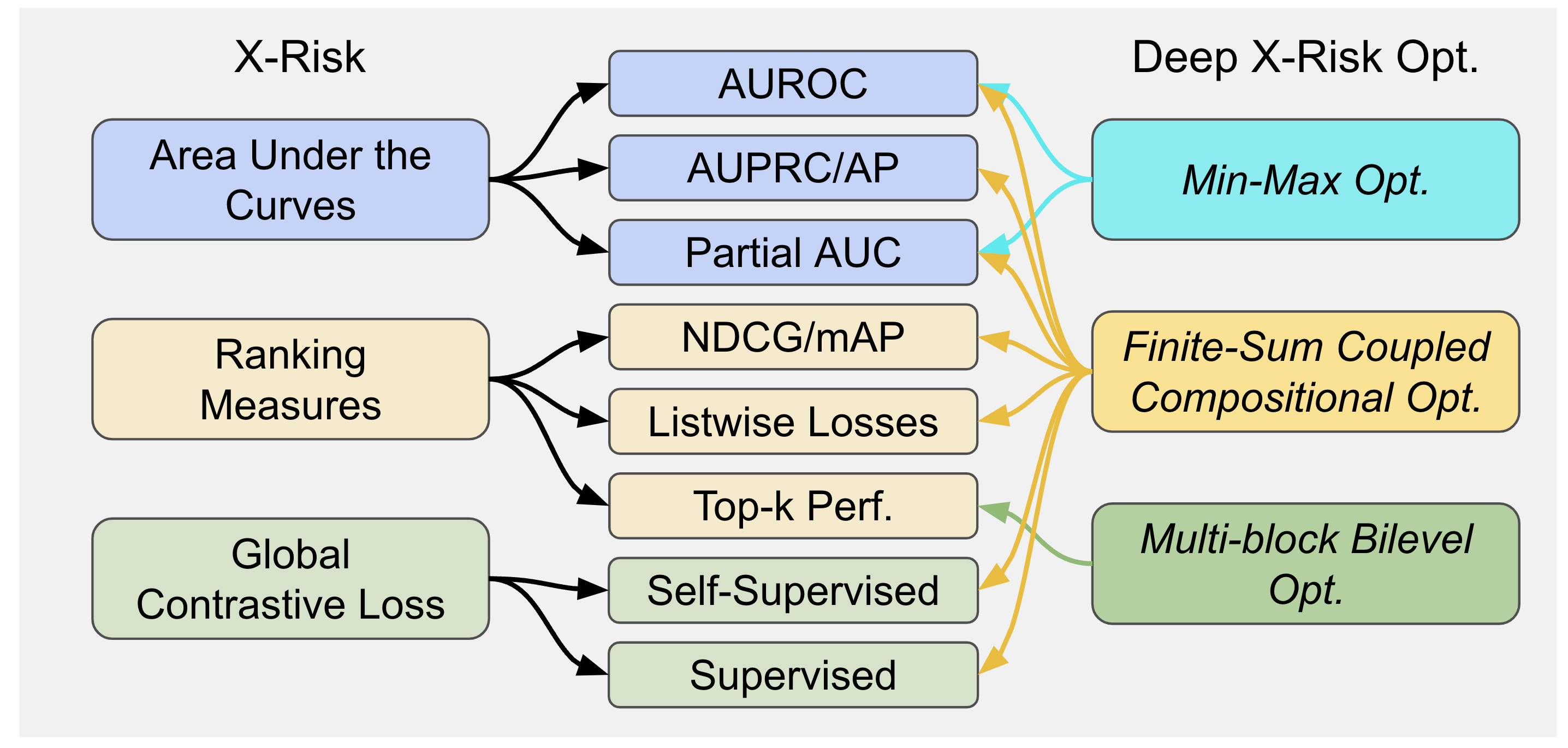}}
\vspace{-0.1in}
\caption{Mappings of  X-risks to optimization problems.}
\label{fig:map}
\end{center}\vspace{-0.15in}
\end{figure}

Besides the common formulation above, in the development of LibAUC library two other optimization problems are also used, including the  min-max optimization and multi-block bilevel optimization. The min-max formulation is used to formulate a family of surrogate losses of AUROC, and the multi-block bilevel optimization is useful for formulating ranking performance measures  defined only on top-$K$ items in the ranked list, including top-$K$ NDCG, precision at a certain recall level, etc. In summary, we present a mapping of different X-risks to different optimization problems in Figure~\ref{fig:map}, which is a simplified one from~\cite{yang2022algorithmic}.

\vspace*{-0.1in}\subsection{X-risks in LibAUC}
Below, we discuss how different X-risks are formulated for developing their optimization algorithms in the LibAUC library.

\textbf{Area Under the ROC Curve (AUROC).} Two formulations have been considered for AUROC maximization in the literature. A standard formulation is the  pairwise loss minimization~\cite{yang2022auc}:
\begin{align*}%\label{eqn:auroc}
\min_{\w\in\R^d} \mathbb{E}_{\x_i\in\S_+}\mathbb{E}_{\x_j\in\S_-}\ell(h_\w(\x_j) - h_\w(\x_i)),
\end{align*}

where $\ell(\cdot)$ is a surrogate loss. Another formulation is following the min-max optimization~\cite{liu2019stochastic,yuan2021large}:

\begin{align*} %\label{eqn:aucminmax}
\min_{\w,a,b}&\max_{\alpha\in\Omega}  \quad \mathbb{E}_{\x_i\sim\S_+}[(h_\w(\x_i)-a)^2] + \mathbb{E}_{\x_j\sim\S_-}[(h_\w(\x_j)-b)^2] \notag\\
 \vspace*{-0.1in}& + \alpha(\mathbb{E}_{\x_j\sim\S_-}[h_\w(\x_j)] -\mathbb{E}_{\x_i\sim\S_+}[h_\w(\x_i)]+c)  - \frac{\alpha^2}{2},
\end{align*}

where $c>0$ is a margin parameter and $\Omega\subset\R$. {\bf In LibAUC,} we have implemented an efficient algorithm (PESG) for optimizing the above min-max AUC margin (AUCM) loss with $\Omega=\R_+$~\cite{yuan2021large}. % We also provided standard optimizers for optimizing different pairwise losses. 
The comparison between optimizing the pairwise loss formulation and the min-max formulation can be found in~\cite{zhu2022benchmarking}.

%  and we consider incorpteing two-way partial
\textbf{Partial Area Under ROC Curve (pAUC)} is defined as area under the ROC Curve with a restriction on the range of false positive rate (FPR) and/or true positive rate (TPR). For simplicity, we only consider pAUC with FPR restricted to be less than $\beta\in(0,1]$. 
 Let $\mathcal{S}^{\downarrow}\left[k_1, k_2\right] \subset \mathcal{S}$ be the subset of examples whose rank in terms of their prediction scores in the descending order are in the range of $\left[k_1, k_2\right]$, where $k_1 \leq k_2$. Then, optimizing pAUC with FPR$\leq\beta$ can be cast into:
\begin{align*}%\label{eqn:epaucd2}
\min_{\w} \frac{1}{n_+}\frac{1}{k}\sum\nolimits_{\x_i\in\S_+}\sum\nolimits_{\x_j\in\S^\downarrow_-[1, k]}\ell(h_\w(\x_j)-h_\w(\x_i)),
\end{align*}
where $k = \lfloor n_-\beta\rfloor$. To tackle challenge of handling $\S^\downarrow_-[1, k]$ for data selection, we consider the following FCCO formulation~\cite{zhu2022auc}: %. The new formulation of optimizing partial AUROC becomes
\begin{align}\label{eqn:pauceskl}
\min_{\w}\frac{1}{n_+}\sum\nolimits_{\x_i\in\S_+}\lambda\log \mathbb{E}_{\x_j\in\S_-}\exp(\frac{\ell(h_\w(\x_j)-h_\w(\x_i))}{\lambda}),
\end{align}
where $\lambda>0$ is a temperature parameter that plays a similar role of $k$. Let $g(\w; \x_i,\S_-)= \E_{x_j\in\S_-} \exp(\ell(h_\w(\x_j)-h_\w(\x_i))/\lambda)$ and $f_i(g)=\lambda \log(g)$. Then (\ref{eqn:pauceskl}) is a special case of FCCO. %The detailed derivation can be found in~\cite{zhu2022auc}.
{\bf In LibAUC,} we have implemented  SOPAs for optimizing the above objective of one-way pAUC with FPR$\leq\beta$ and SOTAs for optimizing a similarly formed surrogate loss of two-way pAUC with FRP$\leq \beta$ and TPR$\geq \alpha$ as proposed in~\cite{zhu2022auc}. %In this paper, we will focus on SOPA-s. 

\textbf{Area Under Precision-Recall Curve (AUPRC)} is an aggregated measure of precision of the model at all recall levels. A non-parametric estimator of AUPRC is  Average Precision (AP)~\cite{boyd2013area}:
%\vspace{-0.1in}

\begin{align*}%\label{eqn:ap}
    \text{AP}&=\frac{1}{n_{+}} \sum\limits_{\x_i\in\S_+} \frac{\sum\limits_{\x_j\in\S_+}\mathbb I(h_\w(\x_j) \geq h_\w(\x_i))}{\sum\limits_{\x_j\in\S} \mathbb I(h_\w(\x_j)\geq h_\w(\x_i))}.
\end{align*}
By using a differentiable surrogate loss $\ell(h_\w(\x_j) - h_\w(\x_i))$ in place of $\mathbb I(h_\w(\x_j)\geq h_\w(\x_i))$, we consider the following FCCO formulation for AP maximization:
\begin{align*}
    \min_{\w} \frac{1}{n_{+}} \sum\limits_{\x_i\in\S_+} f(g_1(\w; \x_i, \S_+), g_2(\w; \x_i, \S)),
\end{align*}
where $g_1(\w; \x_i, \S_+)=\sum_{\x_j\in\S_+}\ell(h_\w(\x_j) - h_\w(\x_i))$, $g_2(\w; \x_i, \S)=\sum_{\x_j\in\S} \ell(h_\w(\x_j)- h_\w(\x_i))$,  and $f(g_1, g_2)=-\frac{g_1}{g_2}$. {\bf In LibAUC}, we implemented the SOAP algorithm with a momentum SGD or Adam-style update~\cite{qi2021stochastic}, which is a special case of SOX analyzed in~\cite{pmlr-v162-wang22ak}. 

\begin{figure*}[t]
\centering
\includegraphics[scale=0.15]{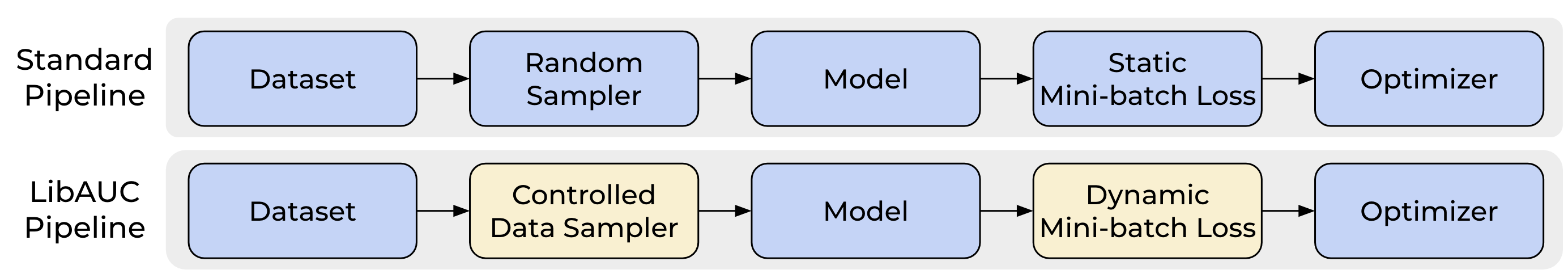}
\vspace{-0.1in}
\caption{The pipeline of \texttt{LibAUC} modules. Highlighted blocks denote the unique modules of the LibAUC library.  } 
\label{fig:libauc_pipeline} \vspace{-0.1in}
\end{figure*}  

\textbf{Normalized Discounted Cumulative Gain (NDCG)} is a ranking performance metric for LTR tasks. The averaged NDCG over all queries can be expressed by  
\begin{align*}%\label{eqn:ndcg}
\quad \frac{1}{N}\sum_{q\in\Q}\frac{1}{Z_q}\sum_{\x_i^q\in \S^+_q} \frac{2^{y^q_i}-1}{\log_2(r(\w; \x^q_i, \S_q)+1) },
\end{align*}
where  $r(\w; \x, \S_q) = \sum_{\x'\in\S_q}\I(h_\w(\x', q) - h_\w(\x, q)\geq 0)$ denotes the rank of $\x$ in the set $\S_q$ respect to $q$, and $Z_q$ is the DCG score of a perfect ranking of items in $\S_q$, which can be pre-computed.  %, and $N$ is the total number of queries. 
%Note that $\x^q_i$ are summed over $\S_q^+$ instead of $\S_q$, because only relevant items have non-zero relevance scores and contribute to NDCG. 
For optimization, the rank function $r(\w; \x_i^q, \S_q)$ is replaced by a differentiable surrogate loss, e.g., $g(\w; \x_i, \S_q)=\sum_{\x'\in\S_q}\ell(h_\w(\x', q) - h_\w(\x, q))$. Hence, NDCG optimization is formulated as FCCO. {\bf In LibAUC}, we implemented the SONG algorithm with a momentum or Adam-style update for NDCG optimization~\cite{qiu2022largescale}, which is a special case of SOX analyzed in~\cite{pmlr-v162-wang22ak}. 

\textbf{Top-$K$ NDCG} only computes the corresponding score for those that are ranked in the top-$K$ positions. We follow~\cite{qiu2022largescale} to formulate top-$K$ NDCG optimization as a multi-block bilevel optimization:% which is given by:
%\vspace{-0.1in}
\begin{align*} 
&\min_{\w}  - \frac{1}{N}\sum_{q=1}^N\frac{1}{Z^K_q}\sum_{\x_i^q\in\S^+_q}\frac{\sigma(h_q(\x_i^q; \w) - \lambda_q(\w))(2^{y^q_i}-1)}{\log_2(g(\w; \x^q_i, \S_q)+1) },\notag\\
 &\lambda_q(\w) = \arg\min_{\lambda} L(\lambda, \w; K, \S_q), \forall q\in\Q,
\end{align*} 
where $\sigma(\cdot)$ is a sigmoid function, $Z^K_q$ is the top-$K$ DCG score of a perfect ranking of items,   and $\lambda_q(\w)$ is an approximation of the $(K+1)$-th largest score of data in the set $\S_q$. The detailed formulation of lower-level problem $L$ can be found in~\cite{qiu2022largescale}. {\bf In LibAUC}, we implemented the K-SONG algorithm with a momentum or Adam-style update for top-$K$ NDCG optimization~\cite{qiu2022largescale}. %, which is a special case of SOX proposed in~\cite{pmlr-v162-wang22ak}.  %The above problem is a special case of MBBO.

\textbf{Listwise CE loss} is defined by a cross-entropy loss between two probabilities of list of scores similar to ListNet in~\cite{cao2007learning}:
\begin{align}\label{eqn:listwise}
  &\min_{\w} -\sum_{q}\sum_{\x^q_i\in\S_q}P(y^q_i)\log \frac{\exp(h_\w(\x^q_i; q)}{\sum_{\x\in\S_q}\exp(h_\w(\x^q_i; q))},
\end{align}
where $P(y^q_i)\propto y^q_i$ denotes a probability for a relevance score $y^q_i$ to be the top one. (\ref{eqn:listwise}) is a special case of FCCO by setting $g(\w; \x^q_i, \S_q) = \E_{\x\in\S_q}\exp(h_\w(\x; q)-h_\w(\x^q_i; q))$ and $f_{q, i}(g) = P(y^q_i)\log (g)$. {\bf In LibAUC}, we implemented an optimization algorithm, similar to SONG, for optimizing listwise CE loss. %, which is a special case of SOX analyzed in~\cite{pmlr-v162-wang22ak}. 

\textbf{Global Contrastive Losses (GCL)} are the global variants of contrastive losses used for unimodal and bimodal SSL. For unimodal SSL, GCL can be formulated as:

\begin{align*}%\label{eqn:gcl} %, \A, \A'\sim \P
   \min_{\w}\E_{\x_i, \x_i^+}\tau\log\E_{\x_j\sim\S_i^-}\exp\left(\frac{h_\w(\x_i)^{\top}h_\w(\x_j) - h_\w(\x_i)^{\top}h_\w(\x_i^+)}{\tau}\right),
\end{align*}
where $\tau>0$ is a temperature parameter and $\x_i^+$ denotes a positive data of $\x_i$. Different from~\cite{simclrv1,clip}, GCL use all possible negative samples $\S_i^-$ for each anchor data  instead of mini-batch samples $\B$~\cite{yuan2022provable}, which helps address the large-batch training challenge in~\cite{simclrv1}. {\bf In LibAUC}, we implemented an optimization algorithm called SogCLR\cite{yuan2022provable} for optimizing both unimodal/bimodal GCL.

As of June 4, 2023, the LibAUC library has been downloaded 36,000 times.  We also implemented two additional algorithms namely MIDAM for solving multi-instance deep AUROC maximization~\cite{DBLP:journals/corr/abs-2305-08040} and iSogCLR~\cite{DBLP:journals/corr/abs-2305-11965} for optimizing GCL with individualized temperature parameters, which are not studied in this paper. 

\begin{figure*}[t]
\begin{minipage}{0.48\textwidth}
\removelatexerror
\begin{algorithm}[H]
\caption{SOPAs for solving pAUCLoss.}
 %Initialize $\w_0$,$\u_0$,$\v_0$,$\eta$, $\beta$, $\gamma$  \;
\For(){$t=0,\ldots,T$}{
    Draw two subsets $\B^t_1\subset \S_+$ and  $\B^t_2\subset\S_-$ \\ %\;
    \For(){$i \in \B^t_1$}{
     {\fboxsep=1pt  \colorbox{columbiablue!40}{ $\u^{t+1}_i =  (1-\gamma)\u^{t}_i  + \gamma g_i(\w_{t};\x_i, \B^t_{2})$  }}\\  
     {\fboxsep=1pt  \colorbox{blond!70}{$p_i^t = \nabla f(\u^{t+1}_i)=\lambda/\u^{t+1}_i$} } %\hfill$\diamond$ theory uses $\u^t_i$}
     }
  Compute the gradient estimator  $G_t$ by
  \hspace*{-0.15in}{\fboxsep=1pt \colorbox{almond!40}{$\frac{1}{|\B^t_1|}\sum_{\x_i\in\B^t_1}\frac{1}{|\B^t_{2}|}\sum_{\x_j\in\B^t_{2}}p_i^t\nabla_\w\exp(\ell(h_{\w_t}(\x_i), h_{\w_t}(\x_j))/\lambda)$}} \\
  %{$\frac{1}{|\B^t_1|}\sum_{\x_i\in\B^t_1}\frac{1}{|\B^t_{2}|}\sum_{\x_j\in\B^t_{2}}p_i^t\nabla_\w\exp(\ell(\frac{h_{\w_t}(\x_i), h_{\w_t}(\x_j)}{\lambda}))$}} \\
  {\fboxsep=1pt \colorbox{babypink!50}{Update the model parameter by an optimizer \hspace{0in}}} \\
 }
\end{algorithm}
\end{minipage}
\begin{minipage}{0.51\textwidth}
\removelatexerror
\begin{algorithm}[H]
\caption{High-level pseudocode for SOPAs.}
 {\fontsize{7.5pt}{5pt}\ttfamily %\selectfont %\ttfamily\footnotesize 
\vspace{0.05in}
%\textcolor{ashgrey}{\# define dynamic loss } \\
\textcolor{hollywoodcerise}{def} pAUCLoss(**kwargs):  \textcolor{ashgrey}{\# dynamic mini-batch loss}\\
\quad\quad sur\_loss = surrogate\_loss(neg\_logits - pos\_logits)   \\
\quad\quad exp\_loss = torch.exp(sur\_loss/Lambda)   \\
\quad\quad {\fboxsep=0pt\colorbox{columbiablue!70}{u[index] = (1 - gamma)*u[index] + gamma*(exp\_loss.mean(1))}}        \\
\quad\quad {\fboxsep=1pt\colorbox{blond!70}{p = (exp\_loss/u[index]).detach()}}        \\
\quad\quad loss = torch.mean(p * sur\_loss)  \\
\quad\quad \textcolor{hollywoodcerise}{return} loss    \\ 
\quad \\
\textcolor{ashgrey}{\# optimization} \\
\textcolor{hollywoodcerise}{for} data, targets, index \textcolor{hollywoodcerise}{in} dataloader:   \\
\quad\quad    logits = model(data)   \\
\quad\quad    {\fboxsep=1pt\colorbox{almond!40}{loss = pAUCLoss(logits, targets, index)}}  \\
\quad\quad    {\fboxsep=1pt\colorbox{almond!40}{optimizer.zero\_grad()\hspace{0.93in}} }  \\
\quad\quad    {\fboxsep=1pt\colorbox{almond!40}{loss.backward()\hspace{1.25in}} }    \\
\quad\quad    {\fboxsep=1pt\colorbox{babypink!50}{optimizer.step()\hspace{0in}}}
}   
\vspace{0.05in}
\end{algorithm}
\end{minipage}
\vspace{-0.1in}
\caption{Left: SOPAs for optimizing pAUC; Right: its pseudo code using automatic differentiation of a dynamic mini-batch loss. The corresponding parts of the algorithm and pseudocode are highlighted in the same color.} 
\label{fig:libauc_algo} \vspace*{-0.1in}
\end{figure*}

\vspace{-0.15in}\section{Library Design of LibAUC}\label{sec:libauc} 
The pipeline of training a DL model in the LibAUC library is shown in Figure~\ref{fig:libauc_pipeline}, which consists of five modules, namely \texttt{Dataset}, \texttt{Data Sampler}, \texttt{Model}, \texttt{Mini-batch Loss}, and \texttt{Optimizer}. The \texttt{Dataset} module allows us to get a training sample which includes its input and output. The \texttt{Data Sampler} module provides tools to sample a mini-batch of examples for training at each iteration. The \texttt{Model} module allows us to define different deep models. The \texttt{Mini-batch Loss} module defines a loss function on the selected mini-batch data for backpropagation. The \texttt{Optimizer} module implements methods for updating the model parameter given the computed gradient from backpropagation. While the \texttt{Dataset, Model,} and \texttt{Optimizer} modules are similar to those in existing libraries, the key differences lie in the \texttt{Mini-batch Loss} and \texttt{Data Sampler} modules. The  \texttt{Mini-batch Loss} module in LibAUC is referred to as \texttt{Dynamic Mini-batch Loss}, which uses dynamically updated variables to adjust the mini-batch loss. The dynamic variables will be defined in the dynamic mini-batch loss, which can be evaluated by forward propagation. In contrast, we refer to the \texttt{Mini-batch Loss} module in existing libraries as \texttt{Static Mini-batch Loss}, which only uses the sampled data to define a min-batch loss in the same way of the objective but on mini-batch data. The\texttt{Data Sampler} module in LibAUC is referred to as \texttt{Controled Data Sampler},  which differ from standard random data samplers in that the ratio of the number of positive data to the number of negative data can be controlled and tuned to boost the performance.  Next, we provide more details of these two and other modules. 

\subsection{Dynamic Mini-batch Loss}  
We first present the stochastic gradient estimator of the objective function, which directly motivates our design of \texttt{Dynamic Mini-batch Loss} module. 

For simplicity of exposure, we will mainly use the FCCO problem of pAUC optimization~(\ref{eqn:pauceskl}) to demonstrate the core ideas of the library design. The designs of other algorithms follow in a similar manner. The key challenge is to estimate the gradient using a mini-batch of samples. To motivate the stochastic gradient estimator, we first consider the full gradient given by 
\begin{align*}
\nabla F(\w) = \E_{\x\in\S_+} \nabla f(g(\w; \x_i, \S_-))\left(\E_{\x_j\in\S_-}\nabla\exp(\ell(\w; \x_i, \x_j)/\lambda)\right).
\end{align*}
To estimate the full gradient, the outer average over all data in $\S_+$ can be estimated by sampling a mini-batch of data $\B_1\subset\S_+$. Similarly, the average over $\x_j\in\S_-$ in parentheses can be also estimated by sampling a mini-batch of data $\B_{2}\subset\S_-$. A technical issue arises when estimating $g(\w; \x_i, \S_-)$ inside $f$. A naive mini-batch approach is to simply estimate $g(\w; \x_i, S_-)$ by using a mini-batch of data in $\B_{2}\subset\S_-$, i.e., $g(\w; \x_i, \B_{2})=\frac{1}{|\B_{2}|}\sum_{\x_j\in\B_{2}}\exp(\ell(\w; \x_i, \x_j)/\lambda)$. However, the problem is that  the resulting estimator $\nabla f(g(\w; \x_i, \B_{2}))$ is biased due to that $f$ is  a non-linear function, whose estimation error will depend on the batch size $|\B_{2}|$. As a result, the algorithm will not converge unless the batch size $|\B_{2}|$ is very large. To address this issue, a moving average estimator is used to estimate $g(\w_t; \x_i, \S_-)$ at the $t$-th iteration~\cite{pmlr-v162-wang22ak,yuan2022provable,qiu2022largescale,qi2021stochastic,zhu2022auc}, which is updated for sampled data $\x_i\in\B^t_1$ according to: 
\begin{align*} %\label{eqn:moving_avg}
&\u^{t+1}_i  =  (1-\gamma)\u^{t}_i  + \gamma g(\w_{t};\x_i, \B^t_{2}) \notag\\
 & =  (1-\gamma)\u^{t}_i  + \gamma \frac{1}{|\B^t_{2}|}\sum_{\x_j\in\B^t_{2}}\exp(\ell(h_{\w_t}(\x_j)- h_{\w_t}(\x_i))/\lambda),
%&\v_{t+1} =\beta_1\v_{t} + (1-\beta_1)\frac{1}{B_1}\sum_{i\in\B^t_1}\nabla f_i({\bf \u^{t}_i}) \nabla g_i(\w_{t};\B^t_{2,i}).\label{eqn:gradient_update}
\end{align*}
where $\gamma\in(0,1)$ is a hyper-parameter.  It has been proved that the averaged estimation error of $\u_i^{t+1}$ for $g(\w_t; \x_i, \S_-)$ is diminishing in the long run.  With the moving average estimators, the gradient of the objective function is estimated by~\footnote{For theoretical analysis $\u_i^{t+1}$ is replaced by $\u^t_i$ in~\cite{pmlr-v162-wang22ak,zhu2022auc}}: 
\begin{align*} 
\hspace*{-0.25in}&G_t = \E_{\x_i\in\B^t_1}\nabla f({\u^{t+1}_i}) \nabla g_i(\w_{t};\x_i, \B^t_{2})\notag\\
&  =\E_{\x_i\in\B^t_1, \x_j\in\B^t_{2}}\nabla f({ \u^{t+1}_i})\nabla_{\w}\exp(\ell(h_{\w_t}(\x_j)- h_{\w_t}(\x_i))/\lambda).\label{eqn:gradient_update}
\end{align*}

The key steps of SOPAs for optimizing pAUC loss are in Algorithm~1~\cite{zhu2022auc}. To facilitate the implementation of computing the gradient estimator $G_t$, we design a dynamic mini-batch loss. 
The motivation of this design is to enable us to simply use the automatic differentiation of PyTorch or TensorFlow for calculating the gradient estimator $G_t$. In particular, on PyTorch we aim to define a loss such that we can directly call \texttt{loss.backward()}  to compute $G_t$. To this end, we define a dynamic variable $p_i=\nabla f(\u_i^{t+1})$ for $\x_i\in\B_1^t$ and then define a dynamic mini-batch loss as $\texttt{loss}=\frac{1}{|\B_1^t|}\sum_{\x_i\in\B_1^t}\frac{1}{|\B_2^t|}\sum_{\x_j\in\B_2^t} p_i\exp(\ell(h_{\w_t}(\x_j) - h_{\w_t}(\x_i))/\lambda)$. However, since $p_i$ depends on $\u_i^{t+1}$ that is computed based on $\w_t$, directly calling \texttt{loss.backward()} for this loss may cause extra differentiation of $p_i$ in term of $\w_t$. To avoid this, we apply the detach operator \texttt{p.detach()} to separate each $p_i$ from the computational graph by returning a new tensor that does not require a gradient. The high-level pseudo code of defining and using the dynamic mini-batch loss for pAUC is given in Algorithm~2, where we use a variable change to define the loss, i.e., $p_i =\nabla f(\u_i^{t+1})\exp(\ell(h_{\w_t}(\x_j) - h_{\w_t}(\x_i))/\lambda)/\lambda$.

Below, we give another example of code snippet to implement the dynamic mini-batch contrastive loss for optimizing GCL.  
\begin{lstlisting}[frame=single, language=Python, numbers=none, label={lst:libauc_dev_template}]
def GCLoss(**kwargs)
    """Defines dynamic mini-batch loss for GCL."""
    # logits: pairwise similarities, labels: pairwise one-hot labels, B: batch size 
    neg_logits = exp(logits/tau) * (1-labels) 
    u = (1-gamma) * u[index] \
      + gamma * sum(neg_logits, dim=1)/(2(B-1))
    p = (neg_logits/u).detach() 
    sum_neg_logits = sum(p * logits, dim=1)/(2(B-1))
    normalized_logits = logits - sum_neg_logits
    loss = -sum(labels * normalized_logits, dim=1) 
    return loss.mean()
\end{lstlisting}
% B = h1.shape[0]  
% labels = cat([one_hot(range(B)), one_hot(range(B))], dim=1)
%  logits = cat([dot(h1, h2.T), dot(h1, h1.T)], dim=1)

\subsection{Controlled Data Sampler}
Unlike traditional ERM, DXO requires sampling  to estimate the outer average and the inner average. In the example of pAUC optimization by SOPAs, we need to sample two mini-batches $\B^t_1\subset\S_+$ and $\B_2^t\subset\S_-$ at each iteration $t$. We notice that this is common for optimizing areas under curves and ranking measures. For some losses/measures (e.g., AUPRC/AP, NDCG, top-$K$ NDCG, Listwise CE), both sampled positive and negative samples will be used for estimating the inner functions. According to our theoretical analysis~\cite{pmlr-v162-wang22ak}, balancing the mini-batch size for outer average and that for the inner average could be beneficial for accelerating convergence.  Hence,  we design a new \texttt{Data Sampler} module to ensure that both positive and negative samples will be sampled and the proportion of positive samples in the mini-batch can be controlled by a hyper-parameter. 

For CID problems, we introduce \texttt{DualSampler}, which takes as input  hyper-parameters such as \texttt{batch\_size} and \texttt{sampling\_{rate}}, to generate the customized  mini-batch samples, where \texttt{sampling\_{rate}} controls the number of positive samples in the mini-batch according to the formula \texttt{\# positives} = \texttt{batch\_size}*\texttt{sampling\_{rate}}. For LTR problems,   we introduce \texttt{TriSampler}, which has  hyper-parameters \texttt{sampled\_tasks} to control the number of sampled queries for backpropogation, \texttt{batch\_size\_per\_task} to adjust mini-batch size for each query, and \texttt{sampling\_rate\_per\_task} to control the ratio of positives in each mini-batch per query. The \texttt{TriSampler} can be also used for multi-label classification problems with many labels such that sampling labels  becomes necessary, which makes the library extendable for our future work. 
\begin{figure}[t]
\centering
\includegraphics[scale=0.28]{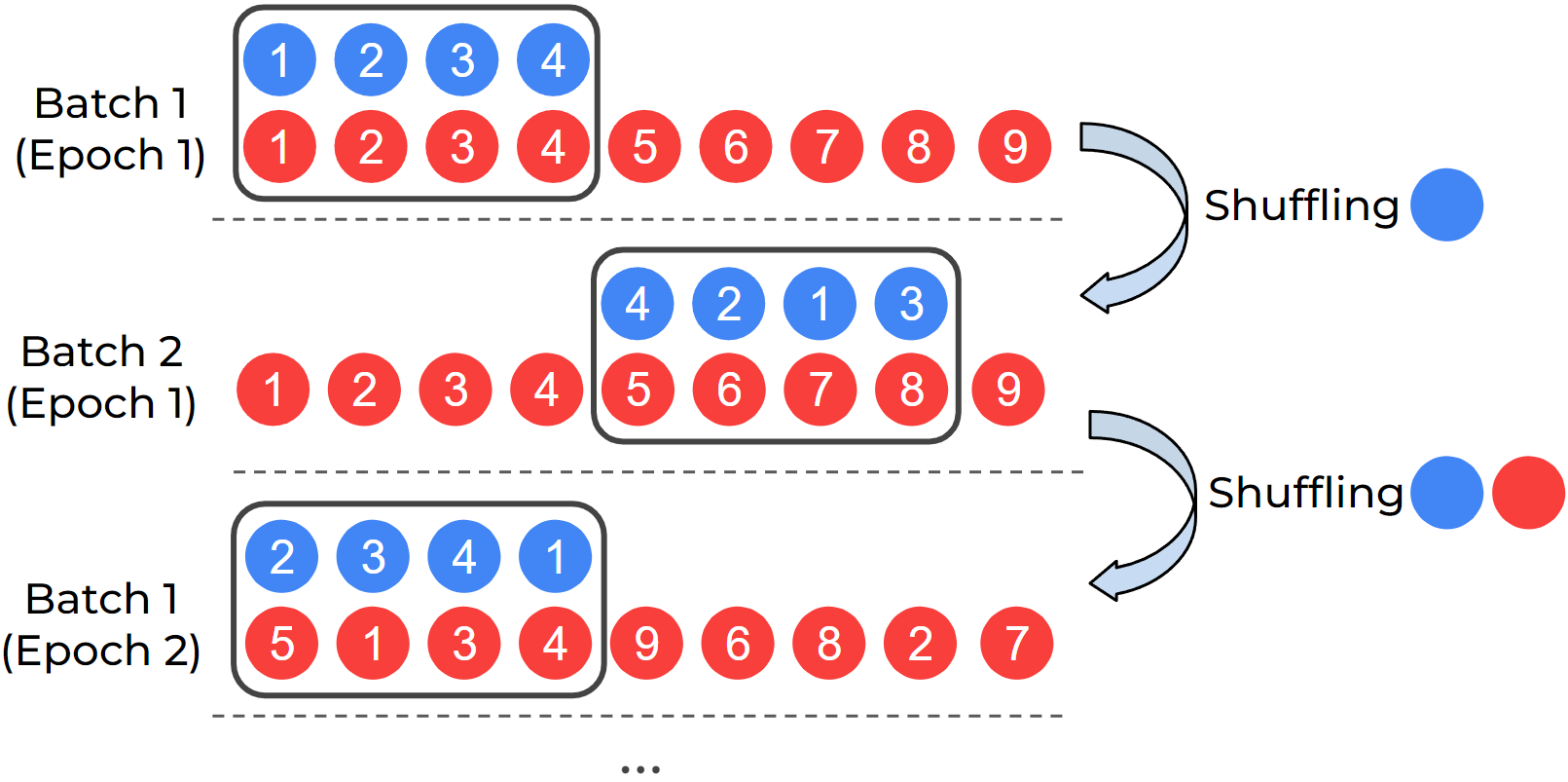}
\vspace*{-0.15in}\caption{Illustration of \texttt{DualSampler} for an imbalanced dataset with 4 positives \textcolor{myblue}{$\bullet$} and 9 negatives \textcolor{myred}{$\bullet$}. } 
\label{fig:data_sampler} 
\end{figure} 
To improve the sampling speed, we have implemented an index-based approach that eliminates the need for computationally intensive operations such as \texttt{concatenation} and \texttt{append}. Figure~\ref{fig:data_sampler} shows an example of \texttt{DualSampler} for constructing mini-batch data with even positive and negative samples on an imbalanced dataset with 4 positives and 9 negatives. We maintain two lists of indices for the positive data and negative data, respectively.  At the beginning, we shuffle the two lists and then take the first 4 positives and 4 negatives to form a mini batch. Once the positive list is used up,  we only reshuffle the positive list and take 4 shuffled positives to pair with next 4 negatives in the negative list as a mini-batch. Once the negative list is used up (an ``epoch" is done),  we re-shuffle both lists and repeat the same process as above. For \texttt{TriSampler}, the main difference is that we first randomly select some queries/labels before sampling the positive and negative data for each query/label. %{\color{red}Note that re-sampling process happens every epoch before training instead only doing it once at the beginning.}  
The following code snippet shows how to define \texttt{DualSampler}  and \texttt{TriSampler}. 
% Additionally, we have pre-computed the \texttt{ideal\_DCG} for L2R problems, which results in a significant speed-up when working with large datasets like MovieLen20M. 

% #Defines DualSampler and TriSampler.
\begin{lstlisting}[frame=single,language=Python, numbers=none, label={lst:libauc_sampler}]
from libauc.sampler import DualSampler, TriSampler
dualsampler = DualSampler(trainSet, 
                          batch_size=32, 
                          sampling_rate=0.1)
trisampler = TriSampler(trainSet, 
                        batch_size_per_task=32,
                        sampled_tasks=5,
                        sampling_rate_per_task=0.1)
\end{lstlisting}

\vspace{-0.1in}\subsection{Optimizer}
With a calculated gradient estimator, the updating rule for the model parameter of different algorithms for DXO follow similarly as (momentum) SGD or Adam~\cite{pmlr-v162-wang22ak,zhu2022auc,qiu2022largescale,yuan2022provable,zhu2022benchmarking,yuan2021large}. Hence, the \texttt{optimizer.step()} is essentially the same as that in existing libraries.  In addition to our built-in optimizer, users can also utilize other popular optimizers from the PyTorch/TensorFlow library, such as Adagrad, AdamW, RMSprop, and RAdam~\cite{duchi2011adaptive,loshchilov2017decoupled,tieleman2012lecture,liu2019variance}. Hence, we provide an optimizer wrapper that allows users to extend and choose appropriate optimizers. For the naming of the optimizer wrapper, we use the name of optimization algorithms corresponding to each specific X-risk for better code readability. An example of the optimizer wrapper for pAUC optimization is given below, where \texttt{mode=`adam'} allows user to use Adam-style update. Another mode  is \texttt{`SGD'}, which takes a momentum parameter as an argument to use the momentum SGD update.

% #An example of optimizer wrapper.
\begin{lstlisting}[frame=single, language=Python, numbers=none, label={lst:libauc_optimizer}]
#An example of optimizer wrapper.
from libauc.optimizers import SOPAs
optimizer = SOPAs(model.parameters(), lr=0.1, mode='adam', weight_decay=1e-4)
\end{lstlisting}

% [emphasize we did some adapations/modifications on the original functions for libauc]
\vspace*{-0.1in}\subsection{Other Modules}
In addition, we provide useful functionalities in other modules, including  \texttt{libauc.datasets}, \texttt{libauc.models}, and \texttt{libauc.metrics}, to help users improve their productivity. The \texttt{libauc.datasets} module provides pre-processing functions for several widely-used datasets, including CIFAR~\cite{krizhevsky2009learning}, CheXpert~\cite{irvin2019chexpert}, and MovieLens~\cite{harper2015movielens}, allowing users to easily adapt these datasets for use with \texttt{LibAUC} in benchmarking experiments. It is important to note that the definition of the Dataset class is slightly different from that in existing libraries. An example is given below, where \texttt{\_\_getitem\_\_} returns a triplet that consists of input data, its label and its corresponding index in the dataset, where the index is returned for accommodating DXO algorithms for updating the $\u^{t+1}_i$ estimators.   The \texttt{libauc.models} module offers a range of pre-defined models for various tasks, including \texttt{ResNet}\cite{he2016deep} and \texttt{DenseNet}\cite{huang2017densely} for classification and \texttt{NeuMF}~\cite{he2017neural} for recommendation. \texttt{libauc.metrics} module offers evaluation wrappers based on \texttt{scikit-learn} for various metrics, such as AUC, AP, pAUC, and NDCG@K. Moreover, it provides an all-in-one wrapper (shown below) to evaluate multiple metrics simultaneously to improve the production efficiency. %Listing~\ref{lst:libauc_evaluator} shows an example of \texttt{evaluator}.

  \begin{lstlisting}[frame=single, language=Python, numbers=none, label={lst:libauc_dataset_def}]
class ImageDataset(torch.utils.data.Dataset):
    """An example of Dataset class"""
    def __init__(self, inputs, targets):
       self.inputs = inputs
       self.targets = targets
    def __len__(self):
        return len(self.inputs)
    def __getitem__(self, index):
        data = self.inputs[index]
        target = self.targets[index]
        return data, target, index 
\end{lstlisting}
\begin{lstlisting}[frame=single, language=Python, numbers=none, label={lst:libauc_evaluator}]
#An evaluator wrapper
from libauc.metrics import evaluator
scores = evaluator(pred,true,metrics=['auc','ap','pauc'])
\end{lstlisting}

\vspace{-0.1in}\subsection{Deployment} %basicstyle=\scriptsize\ttfamily,,  escapechar=`
 Before ending this section, we present a list of different losses, their corresponding data samplers and optimizer wrappers of the LibAUC library in Table~\ref{tab:map_of_loss_and_optimizer}. Finally, we present an example below of building the pipeline for optimizing pAUC using our designed modules. 
\begin{table}[t]
\caption{The list of losses, corresponding samplers and optimizer wrappers in \texttt{libauc}. For a complete list, please refer to the documentation of LibAUC. }
\vspace{-0.1in}
\label{tab:map_of_loss_and_optimizer}
\scalebox{0.75}{
\begin{tabular}{cccc}
\hline
Loss Function & Data Sampler  & Optimizer Wrapper  &  \multirow{2}{*}{Reference}  \\ 
\texttt{libauc.losses}&\texttt{libauc.sampler}&\texttt{libauc.optimizers}&\\\hline
\texttt{AUCMLoss}   & \texttt{DualSampler}   & \texttt{PESG}              & \cite{yuan2021large}   \\  %PDSCA
\texttt{APLoss} & \texttt{DualSampler}  & \texttt{SOAP}                         & \cite{qi2021stochastic}  \\ 
\texttt{pAUCLoss(`1w')} & \texttt{DualSampler}     &\texttt{SOPAs}                                 & \cite{zhu2022auc}  \\ 
\texttt{pAUCLoss(`2w')} & \texttt{DualSampler}     &\texttt{SOTAs}                                 & \cite{zhu2022auc}  \\ 
\hline 
\texttt{NDCGLoss} & \texttt{TriSampler}     & \texttt{SONG}                      & \cite{qiu2022largescale}   \\ 
\texttt{NDCGLoss(topk=5)} & \texttt{TriSampler}     & \texttt{SONG}                      & \cite{qiu2022largescale} \\ 
\texttt{ListwiseCELoss} & \texttt{TriSampler}  & \texttt{SONG}                          &  \cite{qiu2022largescale} \\ 
\hline 
\texttt{GCLoss(`unimodal')} & \texttt{RandomSampler}     & \texttt{SogCLR}         & \cite{yuan2022provable}   \\ 
\texttt{GCLoss(`bimodal')} & \texttt{RandomSampler}     & \texttt{SogCLR}         & \cite{yuan2022provable}   \\ \hline
\end{tabular}}
\end{table}

\begin{lstlisting}[frame=single,language=Python, numbers=none, label={lst:training_pipeline_example}]
#A high-level training pipeline for optimizing pAUC.
from libauc.losses import pAUCLoss 
from libauc.optimizers import SOPAs 
from libauc.sampler import DualSampler
from torch.utils.data import Dataloader
...
dataset = ImageDataset(images, labels)
sampler = DualSampler(dataset,sampling_rate=0.1)
dataloader = DataLoader(dataset, sampler, shuffle=False)
Loss = pAUCLoss('1w') # one-way pAUC loss
optimizer = SOPAs()
...
for data, targets, index in dataloader:
    logits = model(data)
    loss = Loss(logits, targets, index) 
    optimizer.zero_grad()
    loss.backward()
    optimizer.step()
\end{lstlisting}

%[TODO: Show some convergence curve for some experiments]
\section{Experiments}\label{sec:exp}
In this section, we provide extensive experiments on three tasks CID, LTR and CLR. Although individual algorithms have been studied in their original papers for individual tasks, our empirical studies serves as complement to prior studies in that (i) ablation studies of the two unique features for all three tasks provide coherent insights of the library for optimizing different X-risks; (ii) comparison with an existing optimization-oriented library TFCO~\cite{cotter2019two,narasimhan2019optimizing} for optimizing AUPRC is conducted; (iii) a larger scale dataset is used for LTR, and re-implementation of our algorithms for LTR is done on TensorFlow for fair comparison with the TF-Ranking library~\cite{pasumarthi2019tf}; (iv) evaluation of different DXO algorithms based on different areas under the curves is performed exhibiting useful insights for practical use; (v) larger image-text datasets are used for evaluating SogCLR for bimodal SSL.  Another difference from prior works~\cite{qi2021stochastic,zhu2022auc,yuan2021large,qiu2022largescale} is that all experiments for CID and LTR are conducted in an end-to-end training fashion without using a pretraining strategy. However, we did observe the pretraining generally helps improve performance  (cf. the Appendix).

\begin{table*}[t]
\caption{Results on three classification tasks. Best results are marked in bold and second-best results are marked in underline.}
\vspace{-0.1in}
\label{tab:auc_benchmark}
\scalebox{0.85}{
\begin{tabular}{c|ccc|ccc|ccc}
\hline
\multirow{2}{*}{Methods} & \multicolumn{3}{c|}{CIFAR10 (\texttt{imratio}=1\%)} & \multicolumn{3}{c|}{CheXpert (\texttt{imratio}=24.54\%)} & \multicolumn{3}{c}{OGB-HIV (\texttt{imratio}=1.76\%)} \\ \cline{2-10} 
 & \multicolumn{1}{c}{AUROC} & \multicolumn{1}{c}{AP} & pAUC (fpr\textless{}0.3) & \multicolumn{1}{c}{AUROC} & \multicolumn{1}{c}{AP} & pAUC (fpr\textless{}0.3) & \multicolumn{1}{c}{AUROC} & \multicolumn{1}{c}{AP} & pAUC (fpr\textless{}0.3) \\ \hline
CE & \multicolumn{1}{c}{0.687±0.008} & \multicolumn{1}{c}{0.681±0.005} & 0.619±0.003 & \multicolumn{1}{c}{0.853±0.006} & \multicolumn{1}{c}{0.687±0.012} & 0.769±0.011 & \multicolumn{1}{c}{0.765±0.002} & \multicolumn{1}{c}{0.250±0.013} & 0.721±0.004 \\ 
Focal & \multicolumn{1}{c}{0.678±0.006} & \multicolumn{1}{c}{0.671±0.009} & 0.610±0.007 & \multicolumn{1}{c}{0.879±0.004} & \multicolumn{1}{c}{0.737±0.010} & 0.800±0.006 & \multicolumn{1}{c}{0.758±0.004} & \multicolumn{1}{c}{0.241±0.009} & 0.722±0.003 \\
\hline
PESG & \multicolumn{1}{c}{\underline{0.712±0.009}} & \multicolumn{1}{c}{0.706±0.011} & 0.639±0.009 & \multicolumn{1}{c}{\underline{0.890±0.002}} & \multicolumn{1}{c}{\underline{0.759±0.009}} & \underline{0.820±0.003} & \multicolumn{1}{c}{\textbf{0.805±0.009}} & \multicolumn{1}{c}{0.199±0.009} & \underline{0.745±0.007} \\ 
SOAP & \multicolumn{1}{c}{0.711±0.027} & \multicolumn{1}{c}{\textbf{0.717±0.016}} & 0. \textbf{648±0.013} & \multicolumn{1}{c}{0.875±0.048} & \multicolumn{1}{c}{0.757±0.074} & 0.813±0.059 & \multicolumn{1}{c}{0.709±0.008} & \multicolumn{1}{c}{\textbf{0.293±0.004}} & 0.699±0.001 \\ 
SOPAs & \multicolumn{1}{c}{\textbf{0.717±0.005}} & \multicolumn{1}{c}{\underline{0.713±0.002}} & \underline{0.645±0.003} & \multicolumn{1}{c}{\textbf{0.894±0.003}} & \multicolumn{1}{c}{\textbf{0.767±0.008}} & \textbf{0.823+0.006} & \multicolumn{1}{c}{\underline{0.786±0.007}} & \multicolumn{1}{c}{\underline{0.249±0.019}} & \textbf{0.747±0.004} \\ \hline
\end{tabular}}
\vspace*{-0.1in}
\end{table*}

\vspace{-0.05in}\subsection{Classification for Imbalanced  Data}\label{subsec:cid}
% CheXpert:  (12.2+32.2+6.8+31.2+40.3)/5=25.54%
We choose three datasets from different domains, namely CIFAR10 - a natural image dataset  ~\cite{krizhevsky2009learning}, CheXpert - a medical image dataset~\cite{irvin2019chexpert} and OGB-HIV - a molecular graph dataset~\cite{hu2020open}. For CIFAR10, we follow the original paper~\cite{yuan2021large} to construct an imbalanced training set with a positive sample ratio (referred as \texttt{imratio}) of 1\%. For evaluation, we sample 5\% data from training set as validation set and re-train the model using full training set after selecting the parameters and finally report the performance on testing set with balanced positive and negative classes. For CheXpert, we follow the original work~\cite{yuan2021large} by conducting experiments on 5 selected diseases, i.e., Cardiomegaly (\texttt{imratio}=12.2\%), Edema (\texttt{imratio}=32.2\%), Consolidation (\texttt{imratio}=6.8\%), Atelectasis (\texttt{imratio}=31.2\%), Pleural Effusion (\texttt{imratio}=40.3\%), with an average of \texttt{imratio} of 24.54\%. We use the downsized $224 \times 224$ frontal images only for training. Due to the unavailability of testing set, we report the averaged results of 5 tasks on the official validation set. For OGB-HIV, the dataset has an \texttt{imratio} of 1.76\% and we use official train/valid/test split for experiments and report the final performance on testing set. For each setting, we repeat experiments three times using different random seeds and report the final results in \texttt{mean$\pm$std}. 

For modeling, we use ResNet20, DenseNet121, and DeepGCN~\cite{he2016deep,huang2017densely,li2020deepergcn} for the three datasets, respectively. We consider optimizing three losses, namely \texttt{AUCMLoss}, \texttt{APLoss}, \texttt{pAUCLoss} by using PESG, SOAP, SOPAs, respectively. For the latter two, we use the pairwise squared hinge loss with a margin parameter in their definition.  Thus, all losses have a margin parameter, which is tuned in [0.1, 0.3, 0.5, 0.7, 0.9, 1.0]. For \texttt{APLoss} and \texttt{pAUCLoss}, we tune the moving average estimator parameter $\gamma$ in the same range. For \texttt{pAUCLoss}, we also tune the temperature parameter in [0.1, 1.0, 10.0]. For \texttt{DualSampler}, we tune \texttt{sampling\_rate} in [0.1, 0.3, 0.5]. For baselines, we compare two popular loss functions used in the literature, i.e., \texttt{CE} loss and \texttt{Focal} loss. For \texttt{Focal} loss, we tune $\hat{\alpha}$ in [1,2,5] and  $\hat{\gamma}$ in [0.25, 0.5, 0.75]. For optimization, we use the momentum SGD optimizer for all methods with a default momentum parameter 0.9 and tuned initial learning rate in [0.1, 0.05, 0.01]. We decay learning rate by 10 times at 50\% and 75\% of total training iterations. For CIFAR10, we run all methods using a batch size of 128 for 100 epochs. For CheXpert, we train models using a batch size of 32 for 2 epochs. For OGB-HIV, we train models using a batch size of 512 for 100 epochs. To evaluate the performance, we adopt three different metrics, i.e., AUROC, AP, and pAUC (FPR<0.3). We select the best configuration based on the performance metric to be optimized, e.g., using AUROC for model selection of \texttt{AUCMLoss}. The results are summarized in the Table~\ref{tab:auc_benchmark}.

We have several interesting observations. Firstly, directly optimizing performance metrics leads to better performance compared to baseline methods based on ERM framework.  For example, PESG, SOAP, and SOPAs outperform CE and Focal Loss by a large margin in all datasets. This is consistent with prior works. Secondly,  optimizing a specific metric does not necessarily has the best performance for other metrics. For example, on OGB-HIV dataset PESG has the highest AUROC but the lowest AP score, while SOAP has the highest AP score but lowest AUROC and pAUC, and SOPAs has the highest pAUC score. This confirms the importance of choosing appropriate methods in LibAUC for corresponding metrics.  Thirdly, on CheXpert, it seems that optimizing pAUC is more beneficial than optimizing full AUROC. SOPAs achieves better performance than PESG and SOAP in all three metrics. %This is particularly beneficial in many medical applications, as large costs can be incurred due to high False Positive Rate (FPR) and low True Positive Rate (TPR) by a model. 

\begin{figure}[t]
\centering
\includegraphics[scale=0.25]{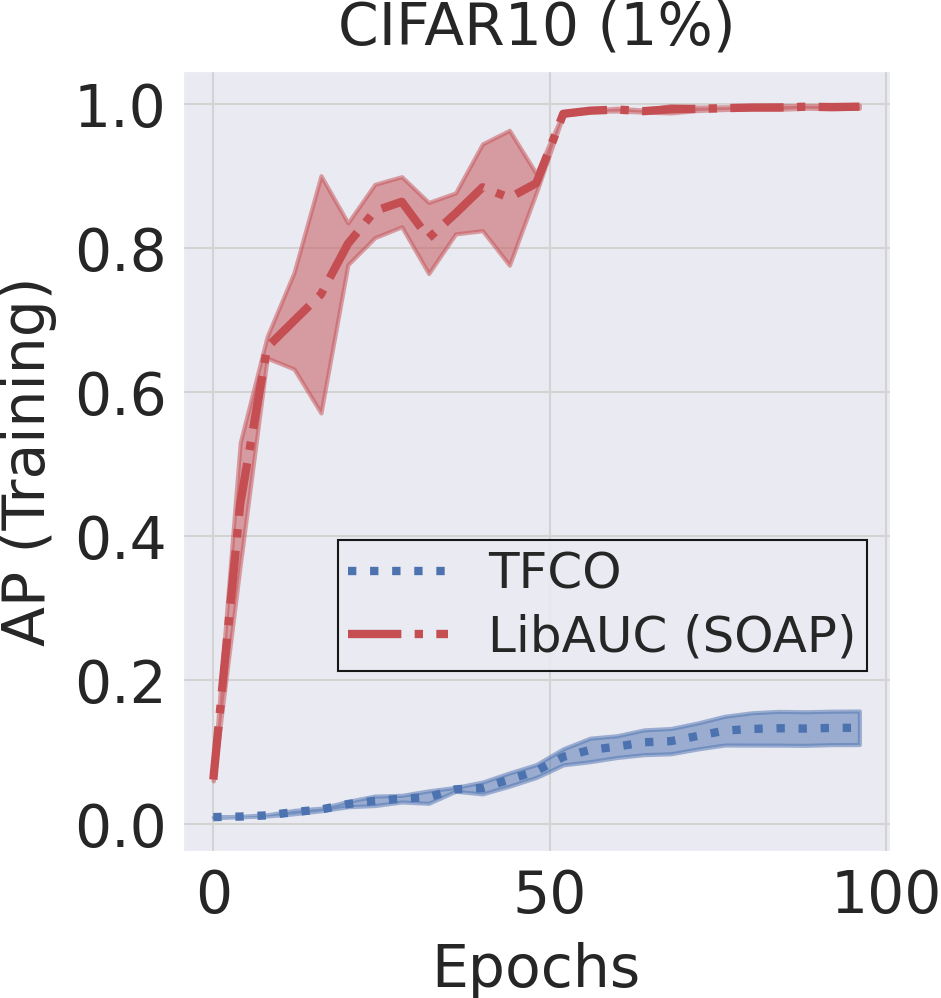}
\includegraphics[scale=0.25]{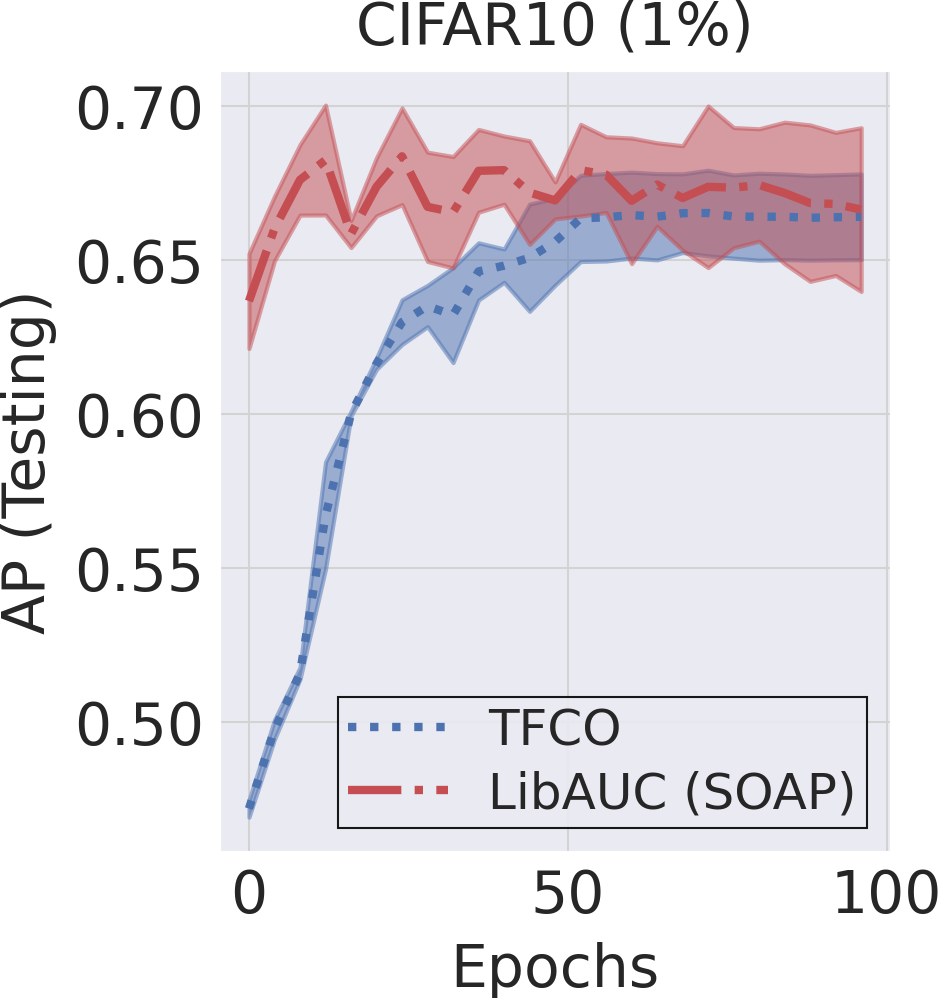}
\includegraphics[scale=0.25]{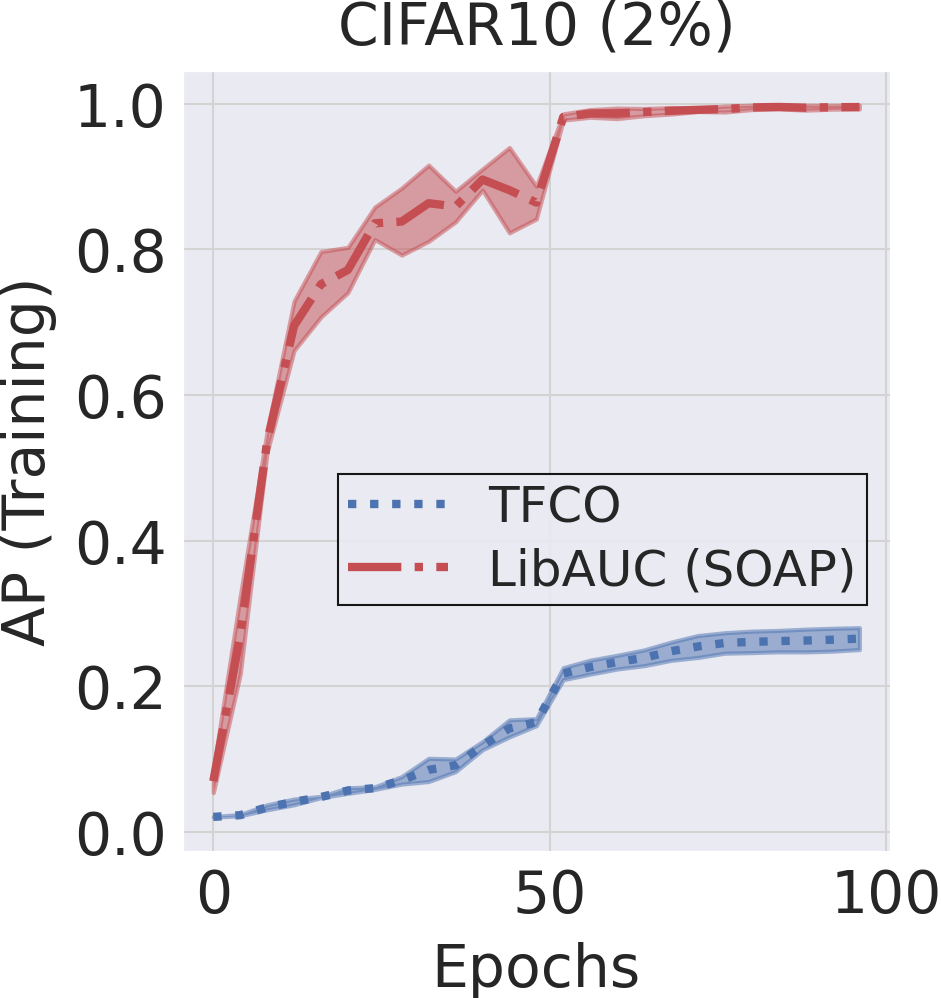}
\includegraphics[scale=0.25]{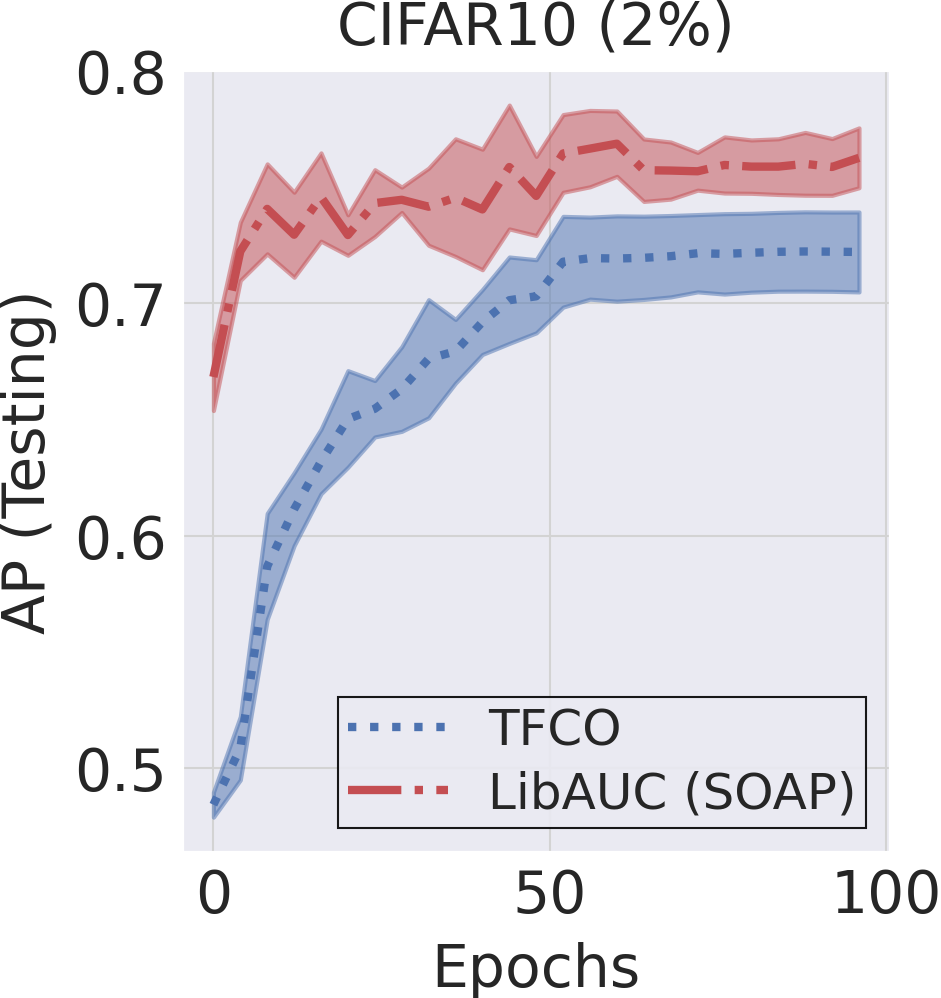}
\vspace{-0.15in}
\caption{Comparison of TFCO and LibAUC.}
\label{fig:tfco} 
\end{figure}

{\bf Comparison with the TFCO library.} We compare LibAUC (SOAP) with TFCO~\cite{cotter2019two,narasimhan2019optimizing} for optimizing AP. We run both methods using batch size of 128 for 100 epochs with Adam optimizer and learning rate of 1e-3 and weight decay of 1e-4 on constructed CIFAR10 with  \texttt{imratio=\{1\%,2\%\}}. We plot the learning curves on training and testing sets  in Figure~\ref{fig:tfco}. The results indicate that LibAUC  consistently performs better than TFCO. 

\begin{figure*}[t]
\begin{minipage}{0.50\textwidth}
\begin{table}[H]
%\caption{Benchmarks on MovieLens datasets.}
\vspace{-0.1in}
\label{tab:exp_ndcg}
\scalebox{0.75}{
\begin{tabular}{cc|cccc}
\hline
&\multirow{2}{*}{Loss} & \multicolumn{2}{c}{MovieLen20M} & \multicolumn{2}{c}{MovieLen25M} \\ \cline{3-6} 
& & \multicolumn{1}{c}{NDCG@5} & NDCG@20 & \multicolumn{1}{c}{NDCG@5} & NDCG@20 \\ \hline
& \texttt{ListMLE} (TF-Ranking)  & \multicolumn{1}{c}{0.2841±0.0007} & 0.3968±0.0004 & \multicolumn{1}{c}{0.3771±0.0003} & 0.4902±0.0003 \\
& \texttt{ApproxNDCG} (TF-Ranking) & \multicolumn{1}{c}{0.3113±0.0001} & 0.4362±0.0001 & \multicolumn{1}{c}{0.3960±0.0003} & 0.5237±0.0001 \\
& \texttt{GumbelNDCG} (TF-Ranking) & \multicolumn{1}{c}{0.3179±0.0003} & 0.4444±0.0001 & \multicolumn{1}{c}{0.4022±0.0002} & 0.5285±0.0013 \\ 
\hline
& \texttt{ListwiseCE} (LibAUC) & \multicolumn{1}{c}{0.3225±0.0005} & 0.4493±0.0003 & \multicolumn{1}{c}{0.4104±0.0001} & \underline{0.5369±0.0001} \\
& \texttt{NDCGLoss(K)}  (LibAUC) & \multicolumn{1}{c}{\underline{0.3325±0.0020}} & \underline{0.4497±0.0037} & \multicolumn{1}{c}{\underline{0.4115±0.0008}} & 0.5249±0.0021\\ 
& \texttt{NDCGLoss} (LibAUC)  & \multicolumn{1}{c}{\textbf{0.3476±0.0001}} & \textbf{0.4769±0.0003} & \multicolumn{1}{c}{\textbf{0.4357±0.0005}} & \textbf{0.5614±0.0003} \\
\hline
%\footnote{Practical version without pretraining. Results with pretraining are included in appendix.}
%& NDCGLoss ($K$) (warmup) & \multicolumn{1}{c}{\underline{xxxx}} & \underline{xxxx} & \multicolumn{1}{c}{\underline{xxx}} & xxx\\ \hline
\end{tabular}}
\end{table}
\end{minipage}
\begin{minipage}{0.45\textwidth}
\begin{figure}[H]
\centering
\includegraphics[scale=0.45]{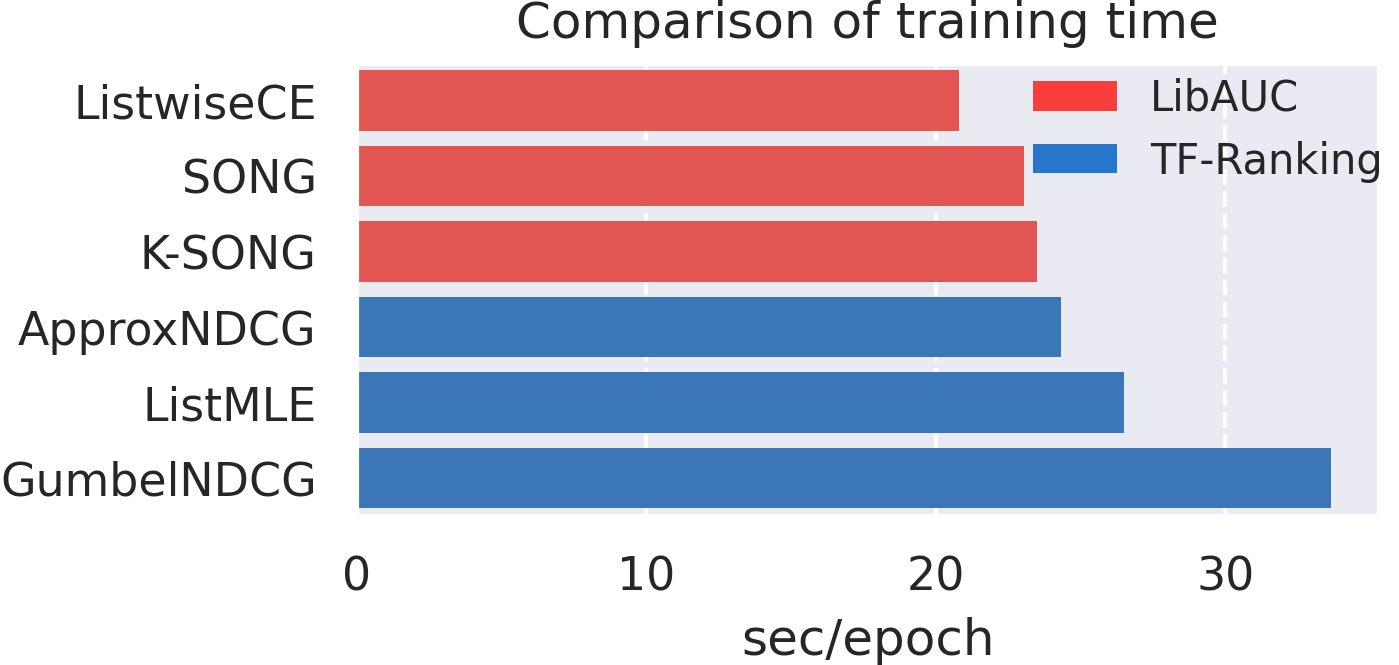}
\vspace{-0.05in}
%\caption{Comparison of training time.}
%\label{fig:result_ndcg} 
\end{figure} 
\end{minipage}
\vspace{-0.1in}
\caption{Left: Results on MovieLens datasets. Right: Comparison of training time for LibAUC and TF-Ranking.} 
\label{fig:ndcg} \vspace*{-0.15in}
\end{figure*}

\vspace{-0.1in}\subsection{Learning to Rank}\label{subsec:ltr}
We evaluate LibAUC on a LTR task for movie recommendation. The goal is to rank movies for users  according to their potential interests of watching based on their historical ratings of movies. We compare the LibAUC library for optimizing \texttt{ListwiseCELoss}, \texttt{NDCGLoss} and  top-$K$ NDCG loss denoted by \texttt{NDCGLoss(K)} against the TF-Ranking library~\cite{pasumarthi2019tf} for optimizing  \texttt{ApproxNDCG}, \texttt{GumbelNDCG}, \texttt{ListMLE},  on two large-scale movie datasets MovieLens20M and MovieLens25M from MovieLens website~\cite{harper2015movielens}.  MovieLens20M contains 20 millions movie ratings from 138,493 users and MovieLens25M contains 25 millions movie ratings from 162,541 users. Each user has at least 20 rated movies. Different from~\cite{qiu2022largescale}, we re-implement the SONG and K-SONG (its practical version) on TensorFlow for optimizing the three losses for a fair comparison of running time with TF-Ranking since it is implemented in TensorFlow. %The TensorFlow versions of SONG/K-SONG have similar prediction performance compared with their PyTorch versions.  
To construct training/validation/testing set, we first sort the ratings based on timestamp for each user from oldest to newest. Then, we put 5 most recent ratings in testing set, and the next 5 most recent items in validation set. For training, at each iteration we randomly sample 256 users, and for each user sample 5 positive items from the remaining rated movies and 300 negatives from all unrated movies. For computing  validation and testing performance, we sample 1000 negative items from the movie list {similar to~\cite{qiu2022largescale}}. %To mitigate computational costs, the number of negative items is limited to 1000. 
%Further studies of the impact of varying the number of positive or negative items are included in Section~\ref{}. This dataset sampling process can be used by calling \texttt{TriSampler} by varying \texttt{num\_pos} and \texttt{num\_neg}. 

For modeling, we use NeuMF~\cite{he2017neural} as backbone network for all methods. We use the Adam optimizer~\cite{kingma2014adam} for all methods with an initial learning rate of 0.001 and weight decay of 1e-7 for 120 epochs by following similar settings in~\cite{qiu2022largescale}. During training, we decrease learning rate at 50\% and 75\% of total iterations by 10 times. For evaluation, we compute and compare NDCG@5 and NDCG@20 for all methods. For \texttt{NDCGLoss}, \texttt{NDCGLoss(K)} and \texttt{ListwiseCELoss}, we tune moving average estimator parameter $\gamma$ in range of [0.1, 0.3, 0.5, 0.7, 0.9, 1.0]. For \texttt{NDCGLoss(K)}, we tune $K$ in [50, 100, 300]. %For \texttt{ApproxNDCG}, \texttt{GumbelNDCG}, \texttt{ListMLE}, we use their recommended parameters. 
We repeat the experiments three times using different random seeds and report the final results in mean$\pm$std. To measure the training efficiency, we conduct the experiments on a NVIDIA V100 GPU  and compute the average training times over 10 epochs. %The final results are summarized in Table~\ref{}.

As shown in the Figure~\ref{fig:ndcg} (left), LibAUC achieves better performance on both datasets. It is worth mentioning that the results of all methods we reported are generally worse than those reported in~\cite{qiu2022largescale}, likely due to different negative items being used for evaluation. In addition, optimizing \texttt{NDCGLoss($K$)} is not as competitive as optimizing \texttt{NDCGLoss}, which is because that we did not use the pretraining strategy used in~\cite{qiu2022largescale}. In Appendix, we show that using pretraining is helpful for boosting the performance of optimizing \texttt{NDCGLoss($K$)}. The runtime comparison, where we report the average runtime in seconds per epoch, is shown in Figure~\ref{fig:ndcg} (right). The results show that our implementation of LibAUC on TensorFlow  is even faster than three methods in {TF-Ranking}. It is interesting to note that LibAUC for optimizing \texttt{ListwiseCE} loss is 1.6$\times$ faster  than TF-Ranking for optimizing \texttt{GumbelLoss} yet has better performance.

\vspace*{-0.05in}\subsection{Contrastive Learning of Representations}\label{subsec:clr}
In this section, we demonstrate the effectiveness of {LibAUC (SogCLR)} for optimizing \texttt{GCLoss} on both uimodal and bimodal SSL tasks. For unimodal SSL, we use two scales of the ImageNet dataset: a small subset of ImageNet with 100 randomly selected classes (about 128k images) denoted as ImageNet-100, and the full version of ImageNet (about 1.2 million images) denoted as ImageNet-1000~\cite{deng2009imagenet}. For bimodal SSL, we use MS-COCO and CC3M~\cite{guo2016ms,sharma2018conceptual} for experiments. MS-COCO is a large-scale image recognition dataset containing over 118,000 images and 80 object categories, and each image is associated with 5 captions describing the objects and their interactions in the image. CC3M is a large-scale image captioning dataset that contains almost 3 million image-caption pairs. For evaluation, we compare the feature quality of pretrained encoder on ImageNet-1000 validation set, which consists of 50,000 images that belong to 1000 classes. For unimodal SSL, we conduct linear evaluation by fine-tuning a new classifier in a supervised fashion after pretraining. For bimodal SSL, we conduct zero-shot evaluation by computing similarity scores between the embeddings of the prompt text and images. Due to the high training cost, we only run each experiment once. It is worth noting that the two bimodal datasets were not used in~\cite{yuan2022provable}.

For unimodal SSL, we follow the same settings in SimCLR~\cite{simclrv1}. We use ResNet-50 with a two-layer non-linear head with a hidden size of 128. We use LARS optimizer~\cite{you2017scaling} with an initial learning rate of \texttt{$0.075\times\sqrt{batch\_size}$} and weight decay of 1e-6. We use a cosine decay strategy to decrease learning rate. We use a batch size of 256 to train ImageNet-1000 for 800 epochs and ImageNet-100 for 400 epochs with a 10-epoch warm-up. For linear evaluation, we train the classifier for additional 90 epochs using the momentum SGD  optimizer with no weight decay. For bimodal SSL, we use a transformer~\cite{radford2019language,vaswani2017attention} as the text encoder (cf appendix for structure parameters) and ResNet-50 as the image encoder~\cite{clip}. Similarly, we use LARS optimizer with the same learning rate strategy and weight decay. We use a batch size of 256  for 30 epochs, with a 3-epoch warm-up. For zero-shot evaluation, we compute the accuracy based on the cosine similarities between image embeddings and text embeddings using 80 different prompt templates similar to~\cite{clip}. Note that we randomly sample one out of five text captions to construct text-image pair for pretraining on MS-COCO. We compare SogCLR with SimCLR for unimodal SSL and with CLIP for bimodal SSL tasks. For {SogCLR}, we tune $\gamma$ in [0.1, 0.3, 0.5, 0.7, 0.8, 0.9, 1.0] and tune temperature $\tau$ in [0.07, 0.1]. %In section~\ref{}, we compare the performance using different $\gamma$. 
All experiments are run on 4-GPU (NVIDIA A40) machines. The results are summarized in Table~\ref{tab:ssl_experiments}.

\begin{table}[t]
% TRY TO TUNE LR A BIT 
\caption{Results for Self-Supervised Learning. Numbers are denoted in \%. SogCLR~\cite{yuan2022provable} is re-implemented in PyTorch.} % [TODO: update imagenet100 by linear eval]
\vspace{-.1in}
\label{tab:ssl_experiments}  
\scalebox{0.8}{
\begin{tabular}{ccccccc}
\hline
\multirow{3}{*}{Dataset} & \multirow{3}{*}{Scale} & \multirow{3}{*}{Modality} & \multicolumn{2}{c}{Acc@1} & \multicolumn{2}{c}{Acc@5} \\ \cline{4-7} 
 &  & &  \multicolumn{1}{c}{\makecell{SimCLR\\ CLIP}} & SogCLR & \multicolumn{1}{c}{\makecell{SimCLR\\ CLIP}} & SogCLR \\  \hline
ImageNet100 &  0.13M &      Image & \multicolumn{1}{c}{78.1} & \textbf{80.3} & \multicolumn{1}{c}{94.9} & \textbf{95.5} \\
ImageNet1000 & 1.2M & Image & \multicolumn{1}{c}{66.5} & \textbf{69.0} & \multicolumn{1}{c}{87.5} & \textbf{89.2} \\ 
\hline
MS-COCO & 0.12M &  Image-Text & \multicolumn{1}{c}{4.6} & \textbf{5.0} & \multicolumn{1}{c}{12.2} & \textbf{12.5} \\
CC3M & 3M &  Image-Text & \multicolumn{1}{c}{19.7} & \textbf{21.4} & \multicolumn{1}{c}{39.3} & \textbf{41.3} \\ \hline
\end{tabular}}
\end{table}

%In particular, SogCLR shows an improvement of 2.2\% and 2.9\% over SimCLR on ImageNet datasets, and an improvement of 0.5\% and 1.6\% over CLIP on two bimodal datasets.
The results demonstrate that SogCLR outperforms SimCLR and CLIP for optimizing mini-batch contrastive losses in both tasks. In particular, SogCLR improves 2.2\%, 2.9\% over SimCLR on ImageNet datasets, and improves 0.5\%, 1.6\% over CLIP on two bimodal datasets. It is notable that the pretraining for ImageNet lasts up to 800 epochs, while the pretraining on the two bimodal datasets is only performed for 30 epochs due to limited computational resources. According to theorems in~\cite{yuan2022provable}, the optimization error of SogCLR will diminish as the training epochs increase. We expect that SogCLR exhibit have larger improvements over CLIP with longer epochs.

\vspace*{-0.1in}
\subsection{Ablation Studies}
In this section, we present more ablation studies to demonstrate the effectiveness of our design and superiority of our library.  

\subsubsection{Effectiveness of Dynamic Mini-batch Losses.}  To verify the effectiveness of the dynamic mini-batch losses, we compare them with conventional static mini-batch losses. To this end, we focus on SOAP, SOPAs, SONG and SogCLR,  and compare their performance with different values of $\gamma$ in our framework. When setting $\gamma=1$, our algorithms will degrade into their conventional mini-batch versions. We directly use the best hyper-parameters tuned in Section~\ref{subsec:cid},~\ref{subsec:ltr} except for $\gamma$, which is tuned from 0.1 to 1.0. The performance is evaluated using AP (SOAP), pAUC (SOPAs), NDCG@5 (SONG), and Top-1 Accuracy (SogCLR), respectively. The final results of this comparison are summarized in Table~\ref{tab:gamma}. Overall, we find that all methods achieve the best performance when $\gamma$ is less than $1$.

%[0.12241724991755092, 0.32190737435022276, 0.06796421448276946, 0.31190878776298636, 0.402555659671146]
\vspace{-0.1in}\subsubsection{Effectiveness of Data Sampler.}  We vary the positive sampling rate (denoted as \texttt{sr}) in the \texttt{DualSampler} for CID by optimizing AUCMLoss, and  in the \texttt{TriSampler} for LTR by optimizing \texttt{NDCGLoss}. For CID, we use three datasets: CIFAR10 (1\%), CheXpert, and OGB-HIV, and tune \texttt{sr}=\{\text{original}, 10\%, 30\%, 50\%\}, where \texttt{sr}=original means that we simply use the random data sampler without any control. Other hyper-parameters are fixed to those found as in Section~\ref{subsec:cid}. The results are evaluated in AUROC and summarized in Table~\ref{tab:sr}. For LTR, we use MovieLens20M dataset. We fix the number of sampled queries (i.e., users) to 256 in each mini-batch and vary the number of positive and negative items, which are tuned in \{1, 5, 10\}  and \{1, 5, 10, 100, 300, 500, 1000\}, respectively. We fix $\gamma=0.1$ and train the model for 120 epochs with the same learning rate, weight decay and learning rate decaying strategies as in section~\ref{subsec:ltr}. The results are evaluated in NDCG@5 and are shown in Table~\ref{tab:sr2}. Both results demonstrate that tuning the positive sampling rate is beneficial for performance improvement.

\begin{table}[t]
\caption{The $\gamma<1$ is better.}
\vspace{-0.1in}
\label{tab:gamma}
\scalebox{0.78}{
\begin{tabular}{cccccccc}
\hline
Method & Dataset & $\gamma=0.1$ & $\gamma=0.3$ & $\gamma=0.5$ & $\gamma=0.7$ & $\gamma=0.9$ & $\gamma=1.0$ \\  \hline
%SOAP & OGB-HIV  & \textbf{0.7166} & 0.7109 & 0.7067 & 0.7039 & 0.6926 & 0.7008 \\ 
SOAP & OGB-HIV  & 0.2745 & 0.2906 & 0.2881 & \textbf{0.2930} & 0.2904 & 0.2864 \\ 
%SOPAs & CIFAR10 & xxx & \textbf{xxx} & xxxx & xxx & xxx & xxx \\ 
SOPAs & OGB-HIV & 0.6404 & 0.7414 & 0.7413 & \textbf{0.7467} & 0.7337 & 0.7383 \\ 
SONG & MovieLens & \textbf{0.3476} & 0.3431 & 0.3384 & 0.3339 & 0.3308 & 0.3290 \\
SogCLR & ImageNet100 & 0.8018 & 0.7956 & \textbf{0.8032} & 0.7974 & 0.7994 & 0.7956 \\ 
SogCLR & CC3M & \textbf{0.2138} & 0.2029 & 0.1931 & 0.1873 & 0.1825 & 0.1778 \\ \hline
\end{tabular}}
\vspace*{0.01in}
%\end{table}
%\begin{table}[t]
\caption{Tuning the sampling rate is beneficial for \texttt{AUCMLoss}.}
\vspace{-0.1in}
\label{tab:sr}
\scalebox{0.8}{
\begin{tabular}{cccccc}
\hline
Dataset & \texttt{imratio} & \texttt{sr}=original & \texttt{sr}=10\% & \texttt{sr}=30\% &  \texttt{sr}=50\% \\ \hline
CIFAR10 & 1\% & 0.7071 & \textbf{0.7124} & 0.7087 & 0.7110 \\ 
Cardiomegaly & 12.2\% & 0.8469 & 0.8515 & \textbf{0.8566} & 0.8378 \\ 
Edema& 32.2\% & 0.9341 & 0.9366 & \textbf{0.9420} & 0.9337 \\
Consolidation & 6.8\% & 0.8888 & \textbf{0.9096} & 0.8832 & 0.8636 \\
Atelectasis & 31.9\% & 0.8231 & 0.8269 & 0.8330 & \textbf{0.8353} \\ 
Pleural Effusion & 40.3\% & 0.9265 & 0.9258 & 0.9249 & \textbf{0.9311} \\
OGB-HIV & 1.8\% & 0.7642 & \textbf{0.8054} & 0.7786 & 0.7752 \\ \hline
\end{tabular}}
%\end{table}
\vspace*{0.015in}
%\begin{table}[t]
\caption{Tuning the sampling rate is beneficial for \texttt{NDCGLoss} on MovieLens20M.}
\vspace{-0.1in}
\label{tab:sr2}
\scalebox{0.85}{
\begin{tabular}{cccccccc}
\hline
Pos/Neg & 1 & 5 & 10 & 100 & 300 & 500 & 1000 \\ \hline
1 & 0.1315 & 0.1617 & 0.1725 & 0.1972 & 0.2039 & 0.2067 & 0.2078 \\
5 & 0.1609 & 0.2289 & 0.2608 & 0.3354 & 0.3480 & 0.3509 & \textbf{0.3522} \\ 
10 & 0.1568 & 0.2083 & 0.2374 & 0.3260 & 0.3417 & 0.3472 & 0.3506 \\ \hline
\end{tabular}}
\end{table}

The results reveal that \texttt{DualSampler} largely boosts the performance for \texttt{AUCMLoss} on CIFAR10 and OGB-HIV when sampling rate (\texttt{sr}) is set to 10\%. It is interesting to note that balancing the data (\texttt{sr}=50\%) did not necessarily improve performance on three cases. However, generally speaking using a sampling ratio higher than the original imbalance ratio is useful. For LTR with \texttt{TriSampler}, we observe a dramatic performance increase when increasing the number of positive samples from 1 to 10, and the number of negative samples from 1 to 300. However, when further increasing the number of negatives from 300 to 1000, the improvement is saturated.

\subsubsection{The Impact of Batch Size.} 
\begin{figure}[t]
\vspace*{-0.05in}\centering
\includegraphics[scale=0.3]{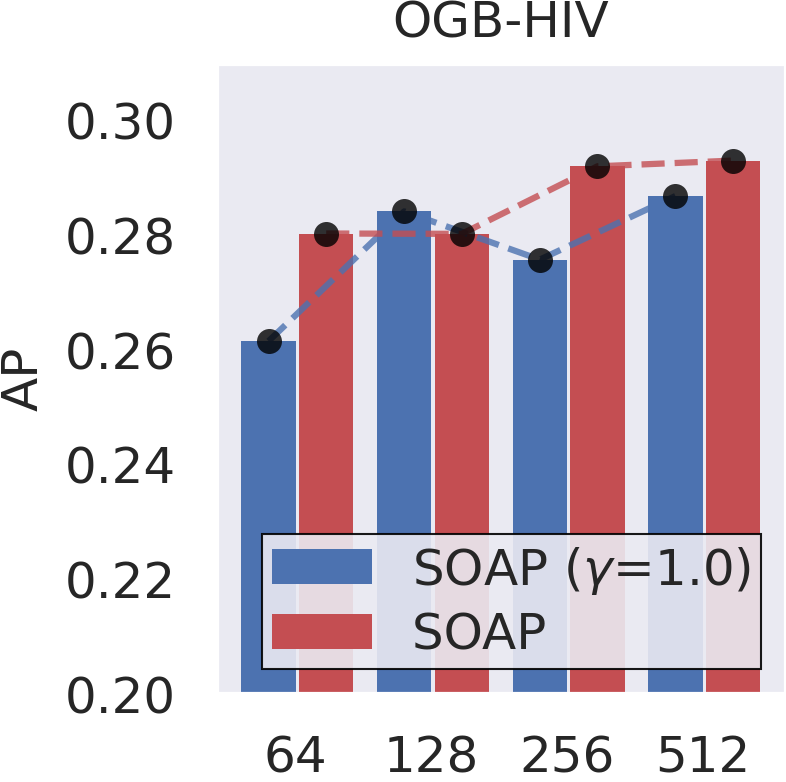}
\includegraphics[scale=0.3]{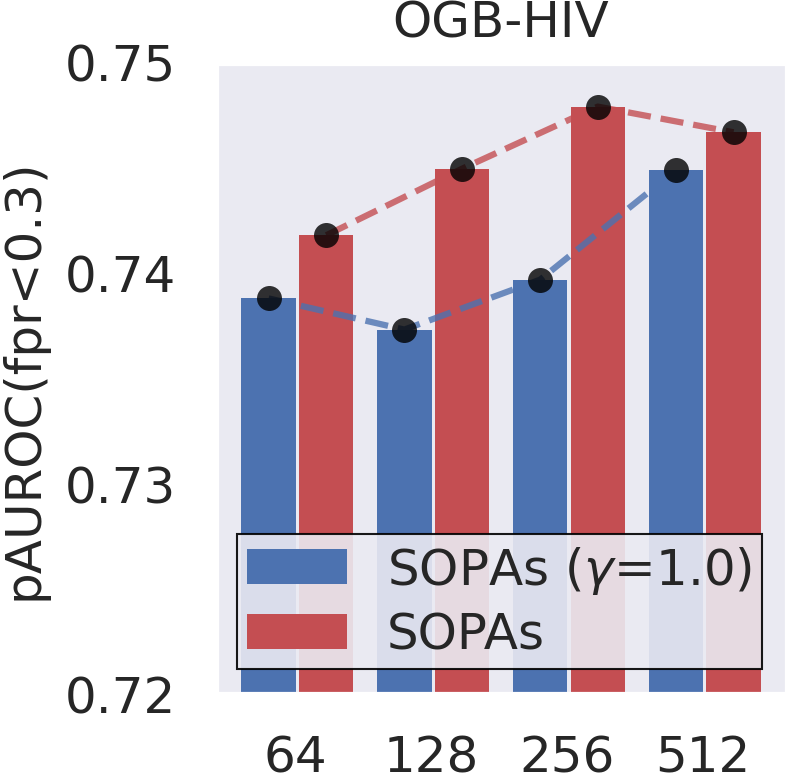} % v2: fixed lambda
\includegraphics[scale=0.3]{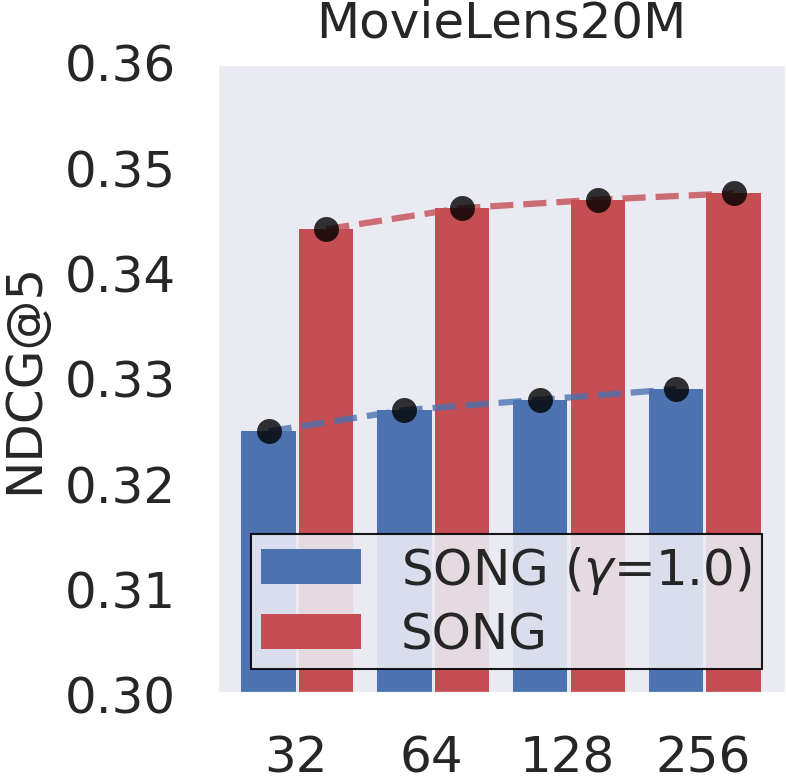}
\includegraphics[scale=0.3]{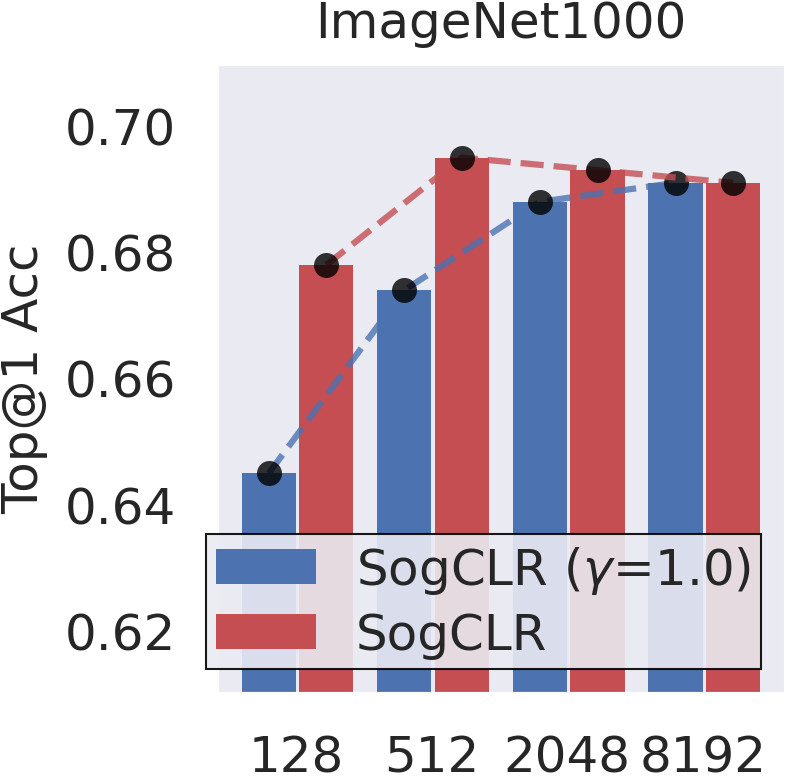}
\vspace{-0.15in}
\caption{Impact of batch size. } 
\label{fig:impact_of_batch_size} 
%\end{figure} 
%\begin{figure}[h]
\vspace{0.05in}\centering
\includegraphics[scale=0.30]{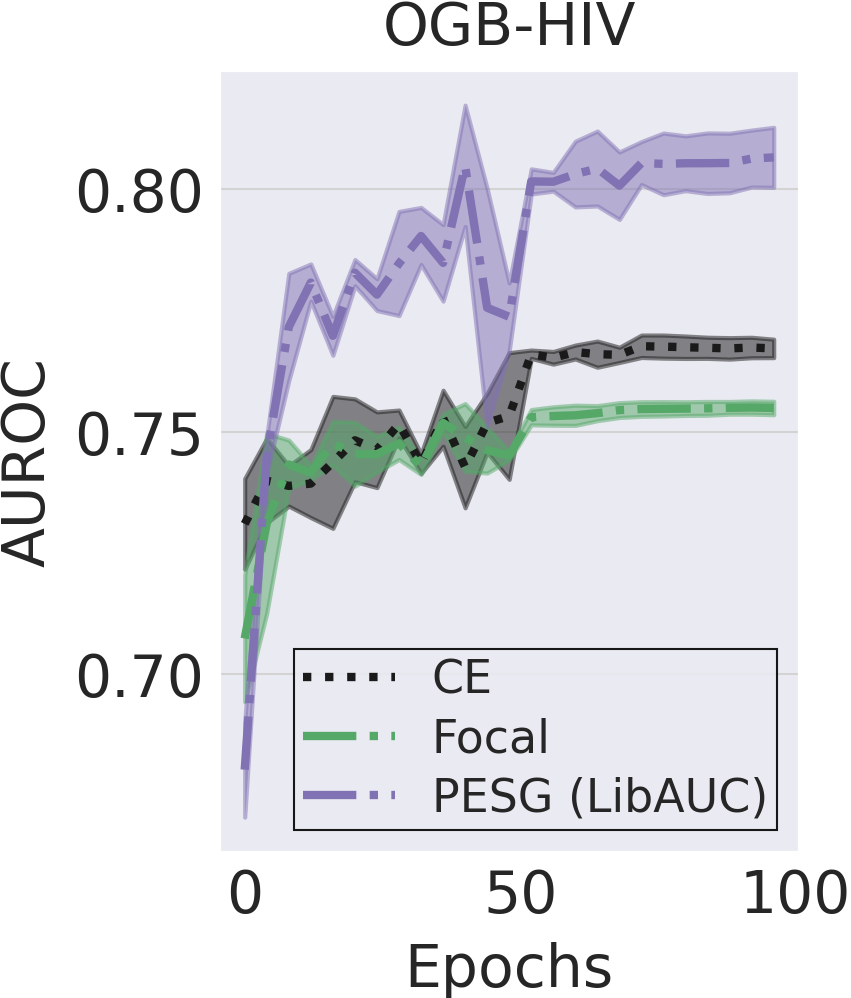}
\includegraphics[scale=0.30]{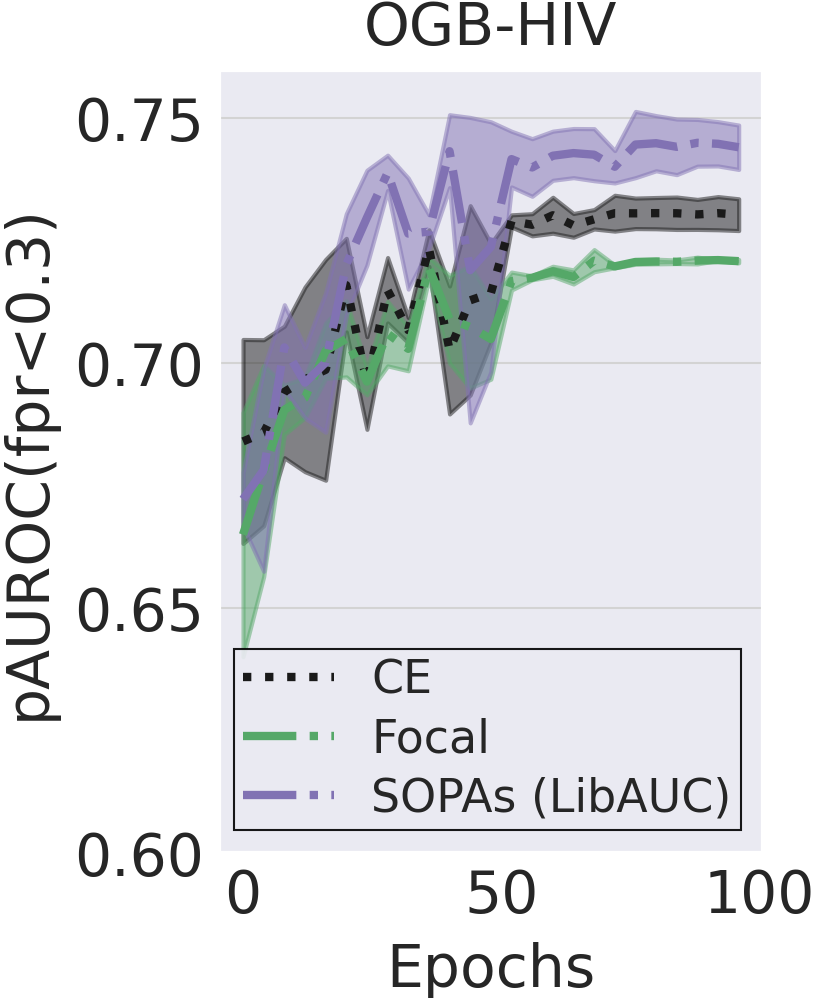}
\includegraphics[scale=0.30]{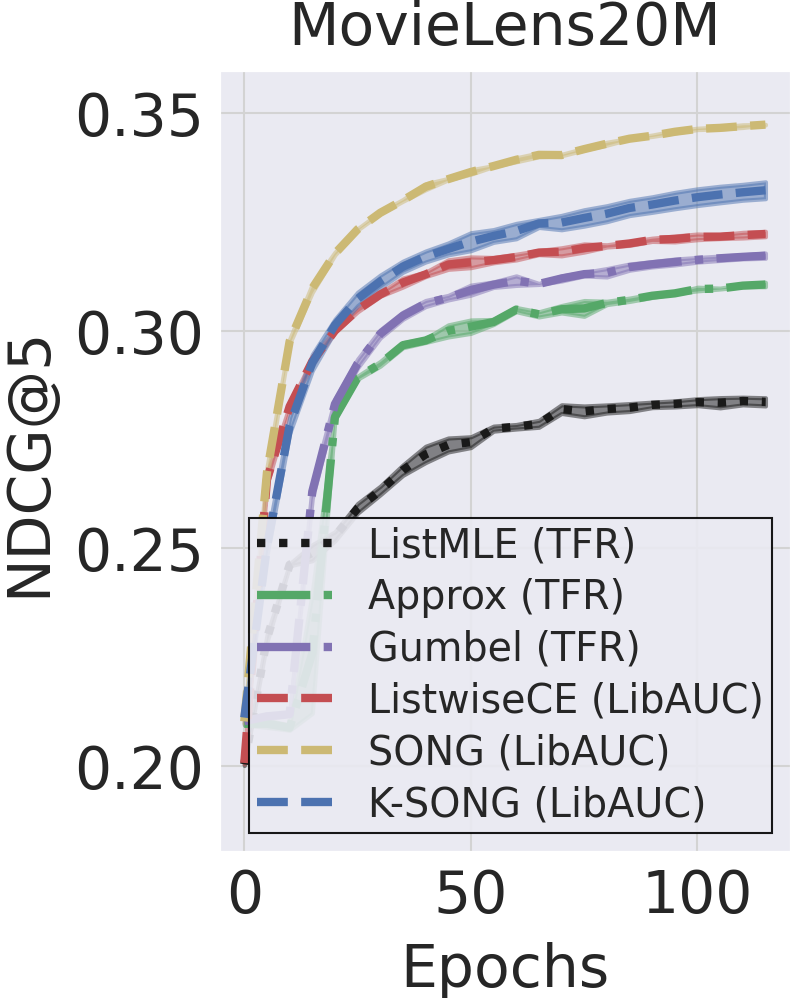}
\includegraphics[scale=0.30]{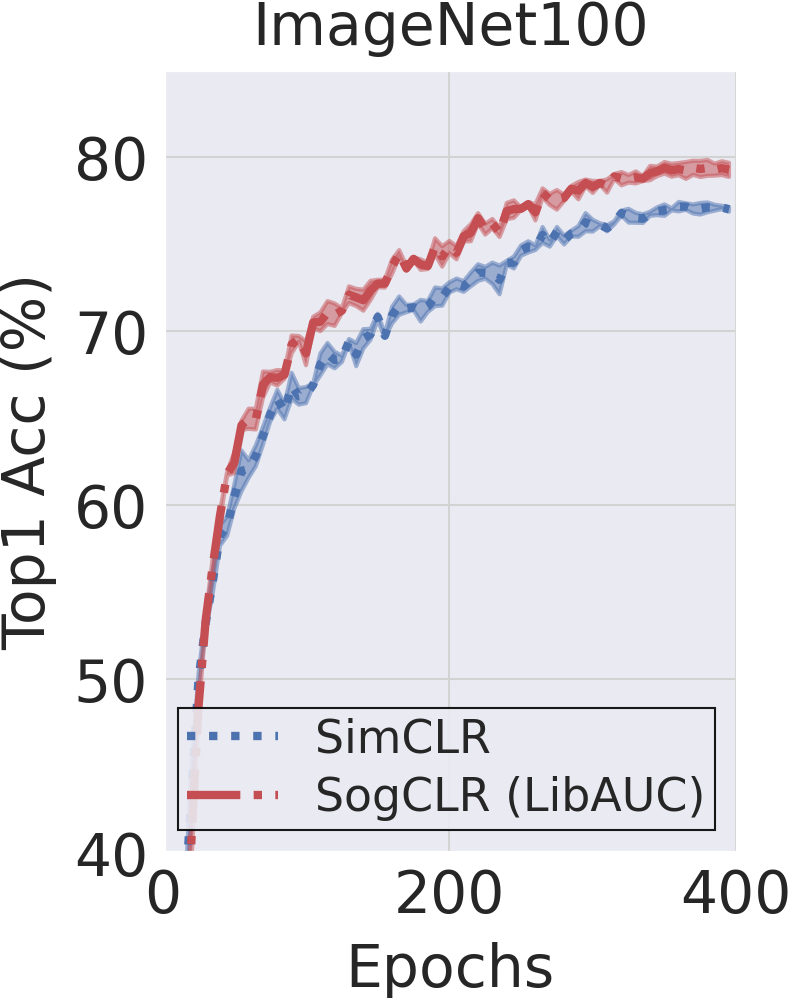}
\vspace{-0.25in}
\caption{Convergence curves of \texttt{LibAUC} algorithms.}
\label{fig:convergence} 
\vspace*{-0.05in}
\end{figure}

We study the impact of the batch sizes on our methods (SOAP, SOPAs, SONG, SogCLR) using dynamic mini-batch losses and that using static mini-batch losses (i.e., $\gamma=1$).  We follow the same experiment settings as in previous section and only vary the batch size. For each batch size, we tune $\gamma$ correspondingly as theories indicate its best value depends on batch size.  For SogCLR, we train ResNet50 on ImageNet1000 for 800 epochs using batch sizes in $\{8192, 2048, 512, 128\}$. For SOAP and SOPAs, we train ResNet20 on OGB-HIV for 100 epochs using batch sizes in $\{512, 256, 128, 64\}$. For SONG, we train NeuMF for 120 epochs on MovieLens20M using batch sizes in  $\{256, 128, 64, 32\}$. The results are shown in Figure~\ref{fig:impact_of_batch_size}, which demonstrates our design is more robust to the mini-batch size.  

\vspace*{-0.1in}\subsubsection{Convergence Speed}  Finally,  we compare the convergence curves of selected algorithms on the OGB-HIV, MovieLens20M, and ImageNet100 datasets. We use the tuned parameters from previous sections to plot the convergence curves on the testing sets. The results are illustrated in Figure~\ref{fig:convergence}. In terms of classification, it is observed that PESG, and SOPAs converge much faster than optimizing CE and Focal loss. For MovieLens20M dataset, we find that SONG has fastest convergence speed compared to all other methods, and K-SONG  (without pretraining) is faster than the other baselines but slower than SONG. In the case of SSL, we observe that SogCLR and SimCLR achieve similar performance at the beginning stage, however, SogCLR gradually outperforms SimCLR as the training time goes longer.

\section{Conclusion \& Future Works}\label{sec:conclusion}
In this paper, we have introduced \textit{LibAUC}, a deep learning library for  X-risk optimization. We presented the design principles of LibAUC and conducted extensive experiments to verify the design principles. Our experiments demonstrate that the LibAUC library is superior to existing libraries/approaches  for solving a variety of tasks including classification for imbalanced data, learning to rank, and contrastive learning of representations. Finally, we note that our current implementation of the LibAUC library is by no means exhaustive. In the future, we plan to implement more algorithms for more X-risks, including performance at the top, such as recall at top-$K$ positions, precision at a certain recall level, etc.

%\begin{acks}
%To Robert, for the bagels and explaining CMYK and color spaces.
%\end{acks}

\bibliographystyle{ACM-Reference-Format}
\balance
\bibliography{paper}

\appendix
%\newpage
\newpage
\section{Appendix}

\iffalse
\subsection{Other formulations}
Here, we present other two key gradient update for MBMMO and MBBO for DXO. The detailed formulation can be referred to~\cite{yang2022algorithmic}. \\

\noindent
\textbf{MBMMO}:
\begin{align} 
&s_{i,t+1} = \Pi_{\Omega}[s_{i,t} + \eta_0 \nabla_s F_i(\w_t, s_{i,t}; \B^t_{2,i})] \\
&\v_{t+1} = \beta_1\v_t  +  (1-\beta_1)\frac{1}{B_1}\sum_{i\in\B_1^t}\nabla_{\w} F_i(\w_{t}, s_{i,t}; \B^t_{2,i}) 
\end{align}

\noindent
\textbf{MBBO}:
\begin{align} 
&\u_i^{t+1}=(1-\gamma_0)  \u_{i}^t+ \gamma_0 g_i(\w_t;\B^t_{2,i})\\
&\lambda_i^{t+1}=\lambda_{i}^t -  \eta_0\nabla_\lambda L_i(\w_t,\lambda_i^t;\bar\B^{t}_{2,i}) \\
&\v_{t+1} = \beta_1\v_t  +  (1-\beta_1)\frac{1}{B_1}\sum_{i\in\B^t_1}\phi_i(\w_t,\lambda_i^t)\nabla g_i(\w_t;\B_{2,i}^t)\nabla f_i(\u_i^t)
\end{align}
\fi

\subsection{Pretraining Strategy} 
\begin{figure}[h]
\centering
\includegraphics[scale=0.3]{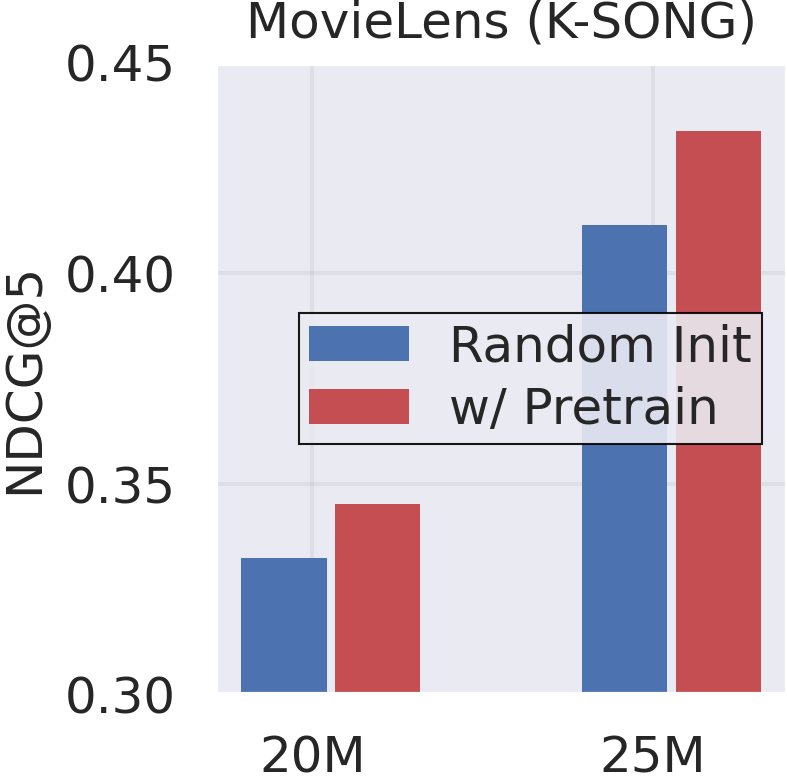}
\includegraphics[scale=0.3]{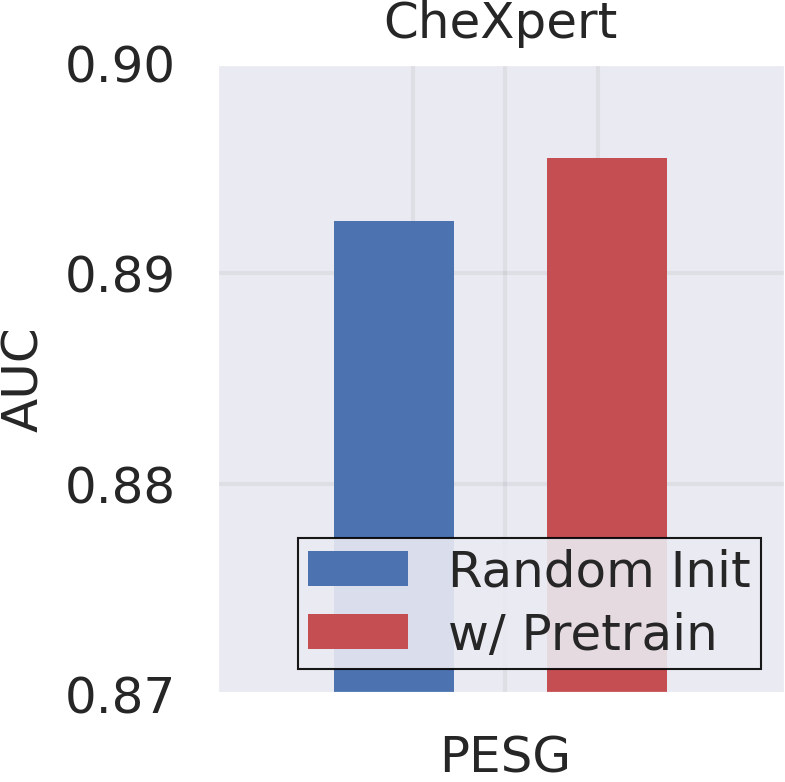}
\includegraphics[scale=0.3]{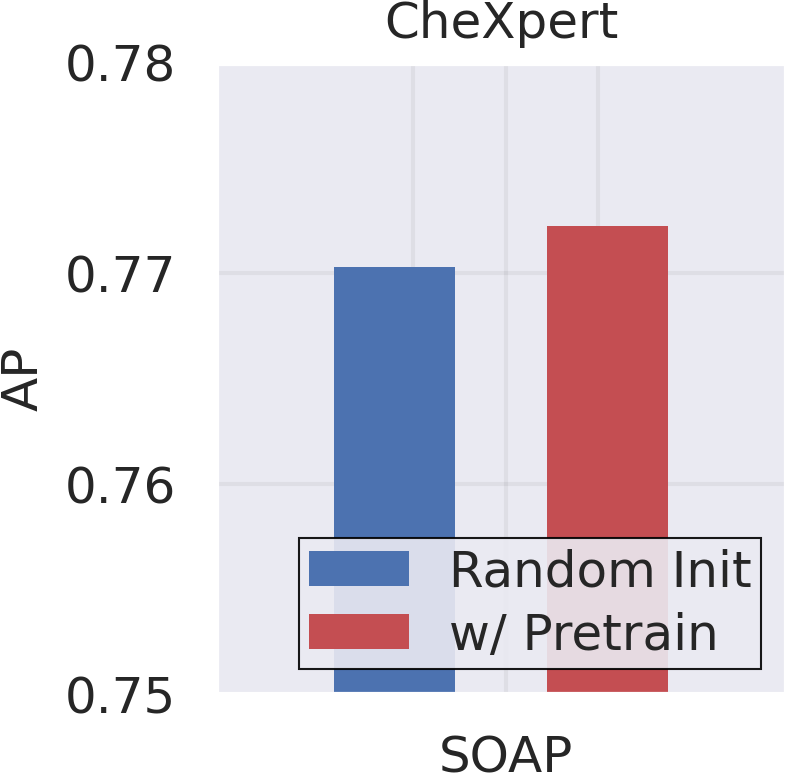}
\includegraphics[scale=0.3]{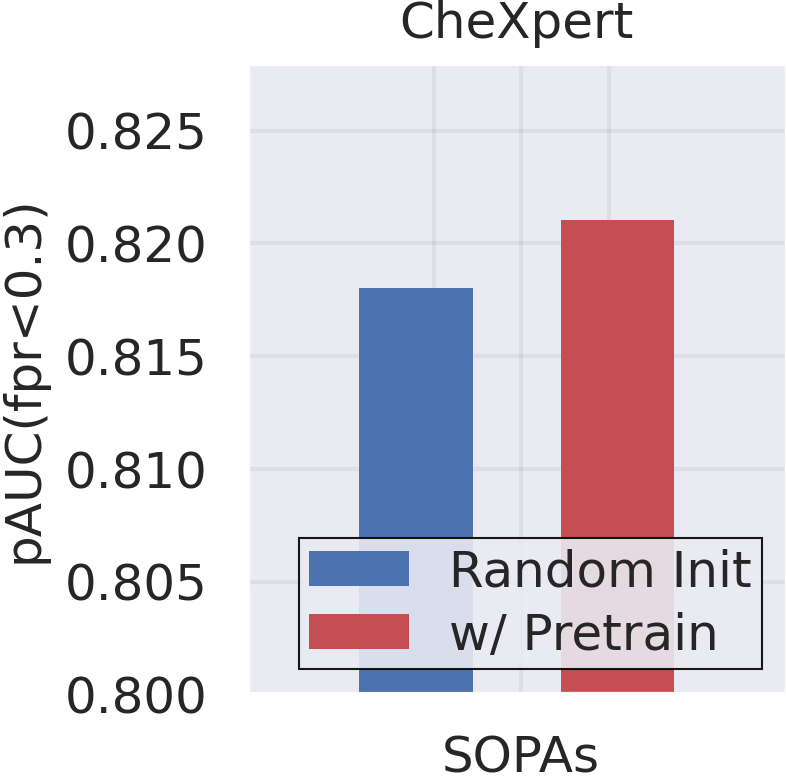}
\vspace{-0.15in}
\caption{Performance comparison of pretraining v.s. random initialization strategies.}
\label{fig:pretrain} 
\end{figure}

We compare the performance of pretraining v.s. random initialization strategy on MovieLens (20M, 25M) with K-SONG and CheXpert with PESG, SOAP, SOPAs, respectively. For K-SONG, we pretrain model using \texttt{ListwiseCELoss} for 30 epochs using learning rate of 0.001 with adam optimizer. Then, we re-initialize last layer and re-train models for 120 epochs by using the tuned parameters in section~\ref{subsec:ltr}. For PESG, SOAP, SOPAs, we use \texttt{CrossEntropyLoss} to pretrain model on CheXpert in multi-label (5 classes) setting for 1 epoch using learning rate of 0.001 with adam optimizer. Then, we re-initialize last layer and re-train models for 2 epochs using the tuned parameters in Section~\ref{subsec:cid} for each individual task. We report average scores of five selected diseases in AUC, AP, pAUC. We present the final results in Figure~\ref{fig:pretrain}. Overall, we can see pretraining boosts the performance of K-SONG by a large margin on two datasets. For CheXpert, we also observe that pretraining can effectively improve the performance on different metrics.

\subsection{Relationship between X-Risk Measures}
\begin{figure}[h]
\centering
\includegraphics[scale=0.17]{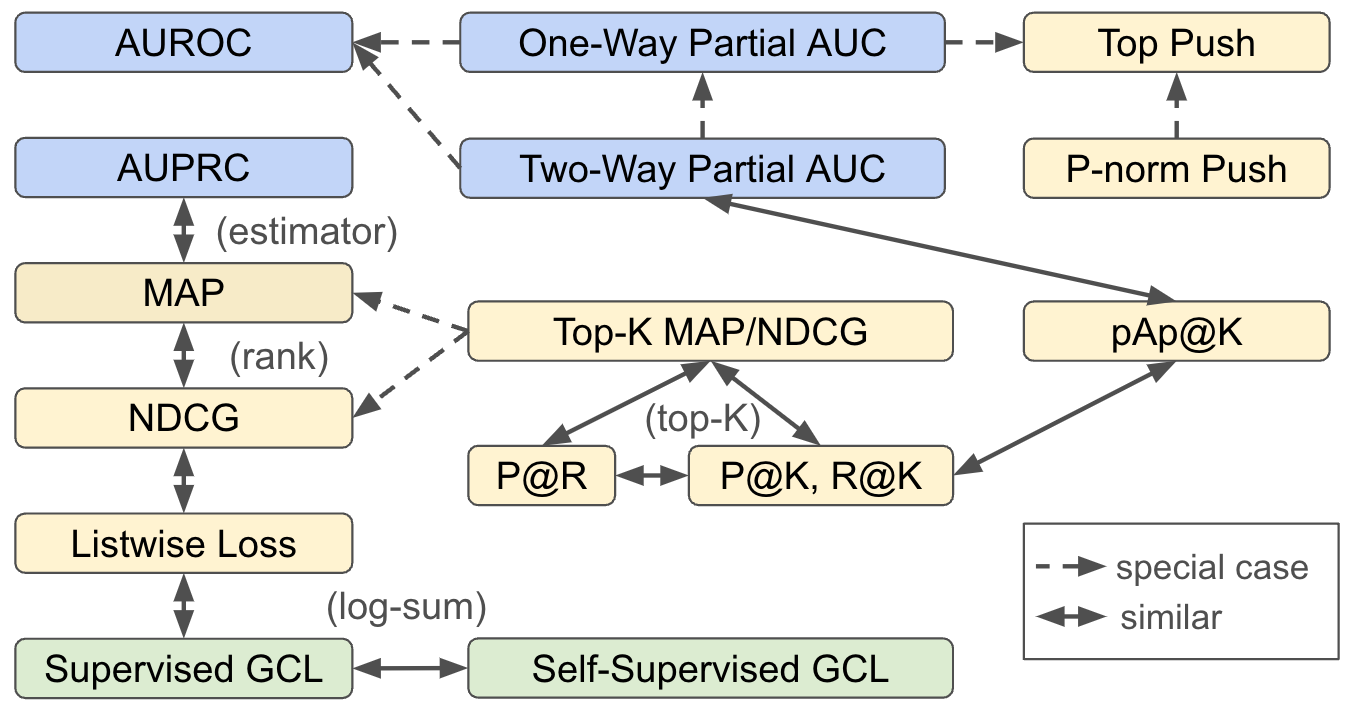}
\caption{Relationships between different X-risks~\cite{yang2022algorithmic}. }
\label{fig:x-risk-relationships} 
\end{figure}

AUROC is a special case of one-way pAUC and two-way pAUC. One-way pAUC with FPR in a range $(0, \alpha)$ is a special case of two-way pAUC. Top Push is a special case of one-way pAUC and p-norm push. AP is a non-parametric estimator of AUPRC. MAP and NDCG are similar in the sense that they are functions of ranks. Top-K MAP, Top-K NDCG, Recall@K (R@K), Precision@K (P@K), pAUC+Precision@K (pAp@K), Precision@Recall (P@R) are similar in the sense that they all involve the computation of K-th largest scores in a set. Listwise losses, supervised contrastive losses, and self-supervised contrastive losses are similar in the sense that they all involve the sum of log-sum term. The above relationships are summarized in Figure~\ref{fig:x-risk-relationships}.

\subsection{Relationship with Stochastic Compositional Optimization Algorithms}
The considered family of problems has a subtle difference from the conventional two level compositional optimization problems studied in the literature (e.g.,~\cite{wang2017stochastic, wang2016accelerating}), though they are closely related. In traditional two-level compositional optimization, the objective is given by $\mathrm E_{\xi}f_\xi(\mathrm E_{\zeta}g_\zeta(\mathbf w))$, where the inner random function $g_\zeta(\mathbf w)$ does not depend on the outer random variable $\xi$. Our problem is given by $\frac{1}{n}\sum_{i=1}^n f_i(g_i(\mathbf w))$, where $g_i(\mathbf w) =\frac{1}{|\mathcal S_i|}\sum_{\mathbf z_j\in \mathcal S_i} \ell(\mathbf w, \mathbf z_i, \mathbf z_j)$, which can be written as $\mathrm E_{i\sim [n]}[f_i(\mathrm E_{\mathbf z_j\sim\mathcal S_i}\ell(\mathbf w, \mathbf z_i, \mathbf z_j))]$. We can see that the key difference between our problem and the conventional two-level compositional optimization problem is that the inner random function $\ell(\mathbf w, \mathbf z_i, \mathbf z_j)$ in our objective not only depends on the inner random variable $\mathbf{z}_j$ but also depends on the outer random variable $\mathbf z_i$. As a result, we cannot simply apply existing algorithms to solving our problems. Instead, we need to maintain and update estimators for all $g_i(\mathbf w) = \mathrm E_{\mathbf z_j\sim\mathcal S_i}\ell(\mathbf w, \mathbf z_i, \mathbf z_j)$ in a random block-wise fashion. Our algorithms were inspired by existing works (e.g.,~\cite{wang2017stochastic, wang2016accelerating,  ghadimi2020single}), with a key difference in that the moving average estimators for $g_i(\mathbf{w})$ are updated only if $\mathbf{z}_i$ is in the sampled mini-batch. 

\subsection{Model Configurations}
For the bimodal pretraining experiments in Section~\ref{subsec:clr}, we implement a small version of CLIP model in PyTorch following the open-source codebase~\cite{open_clip}. The model consists of a modified Transformer and ResNet50~\cite{radford2019language,vaswani2017attention,clip}. The hyperparameters used for building the model are summarized in Table~\ref{tab:clip_config}. For the imbalanced classification on OGB-HIV in Section~\ref{subsec:cid}, we use DeepGCN model~\cite{li2020deepergcn}, which takes inspiration from the concepts of CNNs, e.g., residual connections. We adapt the DeepGCN codebase on OGB-HIV to our experiments, and the hyperparameters used for building the model are summarized in Table~\ref{tab:deep_gcn_config}.

\iffalse
\begin{figure}[h]
\centering
\caption{Configurations for CLIP (Left) and DeepGCN(Right)~\cite{clip,li2020deepergcn}.}
\label{fig:configs}
\begin{minipage}{.5\linewidth}
\centering
\label{tab:clip_config}
\scalebox{0.8}{
\begin{tabular}{|l|c|}
\hline
\multicolumn{1}{|c|}{Hyperparameter} & Value \\ \hline
embed\_dim & 1024 \\ 
image\_resolution & 224$\times$224 \\ 
vision\_layers & {[}3,4,6,3{]} \\ 
vision\_width & 32 \\ 
vision\_patch\_size & null \\ 
context\_length & 77 \\ 
vocab\_size & 49408 \\ 
transformer\_width & 512 \\ 
transformer\_heads & 8 \\ 
transformer\_layers & 12 \\ \hline
\end{tabular}}
\end{minipage}%
\begin{minipage}{.5\linewidth}
\centering
\label{tab:deep_gcn_config}
\scalebox{0.8}{
\begin{tabular}{|l|c|}
\hline
\multicolumn{1}{|c|}{Hyperparameter} & Value \\ \hline
num\_layers & 3 \\
embed\_dim & 256 \\ 
block & res+ \\
gcn\_aggr & max \\
dropout & 0.5 \\ 
temperature & 1.0 \\
norm & batch \\ \hline
\end{tabular}}
\end{minipage}
\end{figure}
\fi 

\begin{table}[H]
\caption{Configuration for CLIP model~\cite{clip}.}
\vspace{-0.1in}
\centering
\label{tab:clip_config}
\scalebox{0.9}{
\begin{tabular}{|l|c|}
\hline
\multicolumn{1}{|c|}{Hyperparameter} & Value \\ \hline
embed\_dim & 1024 \\ 
image\_resolution & 224$\times$224 \\ 
vision\_layers & {[}3,4,6,3{]} \\ 
vision\_width & 32 \\ 
vision\_patch\_size & null \\ 
context\_length & 77 \\ 
vocab\_size & 49408 \\ 
transformer\_width & 512 \\ 
transformer\_heads & 8 \\ 
transformer\_layers & 12 \\ \hline
\end{tabular}}
\end{table}

\begin{table}[H]
\caption{Configuration for DeepGCN model~\cite{li2020deepergcn}.}
\vspace{-0.1in}
\centering
\label{tab:deep_gcn_config}
\scalebox{0.9}{
\begin{tabular}{|l|c|}
\hline
\multicolumn{1}{|c|}{Hyperparameter} & Value \\ \hline
num\_layers & 3 \\
embed\_dim & 256 \\ 
block & res+ \\
gcn\_aggr & max \\
dropout & 0.5 \\ 
temperature & 1.0 \\
norm & batch \\ \hline
\end{tabular}}
\end{table}

\subsection{Additional Experiments}
We run additional experiments to compare our implemented algorithms with two state-of-the-art baselines: (1) \texttt{NeuralNDCG} for LTR~\cite{pobrotyn2021neuralndcg}, which optimizes NDCG by approximating non-continuous sorting operators based on NeuralSort for LRT tasks, and (2) \texttt{VICReg} for CLR tasks~\cite{bardes2022vicreg}, which is based on optimizing invariance, variance, and covariance terms for self-supervised learning of representations. 

For VICReg, we pretrain ResNet-50 with a 2-layer non-linear head with a hidden size of 128 on ImageNet100. We follow the same training parameters as stated in Section~\ref{subsec:clr}. In particular, we pretrain the model for 400 epochs with a batch size of 256, initial learning rate of \texttt{$0.075\times\sqrt{batch\_size}$}, cosine learning rate decay strategy and weight decay of 1e-6. For linear evaluation, we train the classifier for additional 90 epochs using the momentum SGD optimizer with no weight decay. For \texttt{NeuralNDCG}, we train the NeuMF model on MovieLens20M and MovieLens25M datasets. We follow the same training parameters as stated in Section~\ref{subsec:ltr}. In particular, we use the Adam optimizer to train the models with a weight decay of 1e-7 for 120 epochs with the learning rate tuned in the range of [0.001, 0.0005]. For evaluation, we conduct experiments using three different seeds and report the average results in mean±std for NDCG@5 and NDCG@20. The final results for the above two experiments are summarized in Table~\ref{tab:ssl_sota} and Table~\ref{tab:ndcg_sota}.

\begin{table}[h]
\centering
\caption{Comparisons for SSL task on ImageNet100 dataset.}
\vspace{-0.1in}
\scalebox{0.9}{
\begin{tabular}{ccc}
\hline
       & Acc@1 & Acc@5 \\ \hline
VICReg & 74.3     & 92.8     \\
SogCLR & \textbf{80.3}     & \textbf{95.5}     \\ \hline
\end{tabular}}\label{tab:ssl_sota}
\end{table}

\begin{table}[h]
\centering
\caption{Comparisons for LTR task on MovieLens datasets.}
\vspace{-0.1in}
\scalebox{0.9}{
\begin{tabular}{cccc}
\hline
Dataset                      & Methods           & NDCG@5        & NDCG@20       \\ \hline
\multirow{2}{*}{MovieLen20M} & \texttt{NeuralNDCG }      & 0.3181±0.0007 & 0.4424±0.0007 \\ 
                             & \texttt{NDCGLoss} (LibAUC) & \textbf{0.3419±0.0004} & \textbf{0.4709±0.0001} \\ \hline
\multirow{2}{*}{MovieLen25M} & \texttt{NeuralNDCG}       & 0.4059±0.0005 & 0.5322±0.0006 \\ 
                             & \texttt{NDCGLoss} (LibAUC) & \textbf{0.4295±0.0003} & \textbf{0.5566±0.0005} \\ \hline
\end{tabular}}\label{tab:ndcg_sota}
\end{table}

\end{document}